\pdfoutput=1
\documentclass[11pt]{article}
\usepackage[textwidth=400pt,centering]{geometry}
\usepackage{fullpage}
\usepackage[utf8]{inputenc} 
\usepackage[T1]{fontenc}    
\usepackage{url}            
\usepackage{nicefrac}       
\usepackage{xcolor}    
\usepackage{times}

\usepackage{csquotes}
\usepackage{makecell} 
\usepackage{microtype}
\usepackage{graphicx}
\usepackage{subcaption}
\usepackage{booktabs} 
\usepackage[hidelinks]{hyperref}

\usepackage{amsmath, bm, amssymb, mathtools}
\usepackage{physics}

\def\eqref#1{equation~\ref{#1}}

\def\1{\bm{1}}

\DeclareMathAlphabet{\mathsfit}{\encodingdefault}{\sfdefault}{m}{sl}
\SetMathAlphabet{\mathsfit}{bold}{\encodingdefault}{\sfdefault}{bx}{n}

\def\gB{{\mathcal{B}}}

\def\gE{{\mathcal{E}}}
\def\gF{{\mathcal{F}}}

\def\gH{{\mathcal{H}}}

\def\gJ{{\mathcal{J}}}

\def\gO{{\mathcal{O}}}

\def\sN{{\mathbb{N}}}

\newcommand{\E}{\mathbb{E}}

\newcommand{\Reg}{\mathrm{Reg}}

\newcommand{\I}{\mathbbm{1}}

\DeclareMathOperator*{\argmax}{arg\,max}
\DeclareMathOperator*{\argmin}{arg\,min}

\usepackage{amsthm}
\usepackage[capitalize,noabbrev]{cleveref}
\theoremstyle{plain}
\newtheorem{theorem}{Theorem}

\newtheorem{lemma}[theorem]{Lemma}
\newtheorem{corollary}[theorem]{Corollary}
\theoremstyle{definition}

\theoremstyle{remark}
\newtheorem{remark}[theorem]{Remark}

\usepackage{titletoc}
\usepackage{calc}  
\usepackage{array} 

\usepackage[round]{natbib}
\usepackage[nottoc,notlot,notlof]{tocbibind}
\usepackage[vlined, linesnumbered, ruled]{algorithm2e}

\usepackage{pifont}
\usepackage{bbm}
\usepackage{multirow}
\usepackage{braket}
\usepackage{enumitem}
\definecolor{darkblue}{rgb}{0,0.08,0.8}

\usepackage{verbatim}

\usepackage{tikz}
\usetikzlibrary{arrows.meta,positioning}

\tikzset{
  vone/.style={
    circle,
    draw,
    fill=white,
    minimum size=5.5mm,
    inner sep=0pt,
    font=\scriptsize
  },
  vtwo/.style={
    circle,
    draw,
    fill=gray!30,
    minimum size=5.5mm,
    inner sep=0pt,
    font=\scriptsize
  },
  obs/.style={->,>=Stealth,thin},
  bi/.style={<->,>=Stealth,thin}
}

\title{A Parameter-Free Best-of-Both-Worlds Policy for Queueing Bandits}
\author{Jongyeong Lee$^{1}$ \\ \fontsize{10}{12}\selectfont jongyeong@kist.re.kr \and Chansoo Kim$^{1,2}$ \\ \fontsize{10}{12}\selectfont eau@kist.re.kr}
\date{\fontsize{10}{12}\selectfont
$^1$ Korea Institute of Science and Technology 
$^2$ University of Science and Technology
}

\begin{document}
\maketitle

\begin{abstract}
  We study online channel scheduling for a single queue served by $K$ channels under bandit feedback.
  We propose a simple best-of-both-worlds policy that explores when the queue is empty.
  During busy periods, it selects the empirical best channel while the prescribed conditions hold and, if any condition is violated, switches to a robust bandit policy to clear the queue.
  For arbitrary arrivals and services determined by an oblivious adversary, our policy achieves $\tilde{\gO}(\sqrt{K L_T})$ worst-case queue length regret with high probability, where $L_T$ is the longest busy-period length.
  This result improves the previous $T^{3/4}$ dependence and matches the worst-case lower bound up to logarithmic factors.
  Under bounded i.i.d.~stochastic models with a unique best stabilizing channel, we show that the cumulative queue length regret converges to a finite problem-dependent constant, and further establish the logarithmic expected worst-case queue length regret~bound.
\end{abstract}

\section{Introduction}\label{sec: intro}
Bandit problems provide a fundamental framework for sequential decision-making with partial feedback~\citep{lattimore2019bandit}.
In stochastic environments, the learner seeks to identify arms with high expected rewards, whereas adversarial environments require robustness without distributional assumptions.
Best-of-both-worlds policies adapt to both regimes within a single policy~\citep{bubeck2012best}.
For classical bandits, such policies achieve logarithmic, problem-dependent regret in stochastic environments while retaining near-optimal guarantee in adversarial environments without knowing the regime in advance~\citep{auer2016algorithm, zimmert2021tsallis, honda2023follow}.
A natural question is whether such adaptation is possible when decisions have \emph{lasting} effects on a system and performance depends on the system states induced by past decisions rather than solely on the sum of immediate outcomes.

An important example of such state-dependent decision-making arises in queueing bandits, which model scheduling systems with unknown service outcomes~\citep{krishnasamy2016regret,krishnasamy2021learning}.
In this problem, jobs arrive sequentially, and in each round, the scheduler selects one of the channels to serve the queue and observes only the service provided by the selected one.
A poor selection may leave more work unserved in the system, inducing additional backlog in subsequent rounds, whereas the accumulated effect of earlier mistakes disappears once the queue becomes empty.
Therefore, standard bandit regret does not fully characterize queueing performance since the timing of service losses and the regenerative structure of the queue are both important.

\begin{table*}[t]
\renewcommand{\arraystretch}{1.3}
\caption{Comparison of queue length regret guarantees. Here, (S) indicates that the unique best arm is stabilizing, while (E) and (P) indicate guarantees in expectation and with high-probability, respectively. The quantity $L_T$ denotes the length of the longest busy period and satisfies $L_T\leq T$. The queue dynamics considered in \citet{stahlbuhk2021learning}, given in~(\ref{def: queue stan}), differ from those considered in \citet{krishnakumar2025minimizing} and ours, given in~(\ref{def: queue lindley}).} \label{tab: regret}
\vspace{-0.5em}
\begin{center}
\resizebox{\textwidth}{!}{\begin{tabular}{llll}
\textbf{Metric} & \textbf{Reference}  &\textbf{Assumptions} & \textbf{Regret} \\
\midrule
\multirow{3}{*}{\makecell[l]{$ \max\limits_{t\in[T]} \max\limits_{i\in[K]} \qty(Q_t^\pi - Q_t^i)$\\ (worst-case regret)}}
 & \citet{krishnakumar2025minimizing} & oblivious $A_t\geq0$, $R_t \in [0,1]^K$ & $\tilde{\gO}(\sqrt{K}T^{3/4})$ (P)\\
 & Algorithm~\ref{alg: policy} (Theorem~\ref{thm: adv}) & oblivious $A_t\geq0$, $R_t \in [0,1]^K$ & $\tilde{\gO}(\sqrt{KL_T})$ (P)\\
& Algorithm~\ref{alg: policy} (Corollary~\ref{cor}) & i.i.d.~bounded $A_t, R_t$, (S), (\ref{asm: lb on rho}) & $\gO(\log(KT)/\Delta_{\min})$ (E)\\
\midrule
\multirow{2}{*}{\makecell[l]{$\sum_t \E[ Q_t^\pi-Q_t^{i^*}]$\\ (cumulative regret)}}
 & \citet{stahlbuhk2021learning} & i.i.d.~Bernoulli $A_t,R_t$, (S)  & $\gO(1)$\\
 & Algorithm~\ref{alg: policy} (Theorem~\ref{thm: sto cum}) & i.i.d.~bounded $A_t,R_t$, (S), (\ref{asm: lb on rho}) & $\gO(1)$\\
\midrule
\end{tabular}}
\end{center}
\vspace{-1em}
\end{table*}

A classical objective in queueing systems is stability~\citep{dai1995positive}.
One commonly used notion is mean-rate stability, which requires $\lim_{T\to \infty}\E[Q_T^\pi]/T=0$, where $Q_T^\pi$ is the queue length under the policy $\pi$ at round $T$~\citep{neely2010stochastic}.
When service statistics are unknown, one should learn which channels are effective while controlling the queue lengths~\citep{yang2023learning,freund2024efficient}.
Stability alone, however, does not quantify the additional backlog incurred by learning, since two policies may both stabilize the system while producing substantially different queue trajectories.

This limitation motivates finer performance metrics for queueing bandits that compare learning policies with an informed oracle.
For the single queue model, \citet{krishnasamy2016regret} introduce the queue length regret $\E[Q_{T}^\pi - Q_{T}^{*}]$ where the oracle knows the service statistics and always selects the best channel.
They show that this quantity initially behaves similarly to classical bandit regret, but eventually decreases at rate $\tilde{\gO}(1/T)$.
Subsequent work extends this analysis to multiple queues~\citep{krishnasamy2021learning,kim2024queueing} and contextual systems~\citep{bae2026queue}.

Another criterion is the cumulative queue length regret, $\sum_t \E[Q_{t}^\pi - Q_{t}^{*}]$, which measures the backlog accumulated over the horizon.
For a stochastic single queue system with at least one stabilizing channel, \citet{stahlbuhk2021learning} construct queue-dependent policies with $\gO(1)$ cumulative regret.
This is fundamentally stronger than summing a queue length regret of order $1/t$, which would yield logarithmic cumulative regret.
Complementarily, \citet{freund2023quantifying} study the maximum increase in expected time-averaged queue length.

These results rely mainly on stochastic assumptions on arrivals and services, and much less is known when both processes can vary arbitrarily.
In the oblivious adversarial regime, \citet{krishnakumar2025minimizing} study the largest queue length regret over the horizon relative to the best constant policy.
For a single queue served by $K$ channels, they obtain a high-probability bound of $\tilde{\gO}(\sqrt{K}T^{3/4})$.
Since a reduction from adversarial bandits gives a lower bound $\Omega(\sqrt{KT})$, this result leaves a gap in the dependence on $T$.
Moreover, an adversarially robust policy is particularly desirable if it retains the stronger guarantees available in stochastic regimes.
Therefore, it is desirable to design a single policy that achieves near-optimal worst-case queue length regret in adversarial regimes while retaining constant expected cumulative regret in stochastic regimes. 

The main difficulty is that queue length regret is governed by service losses accumulated over the \emph{current} busy period, whose endpoints depend on the policy itself and may therefore be random.
Hence, a bandit regret measured over the entire horizon can be too coarse, and the analysis must instead control regret over data-dependent intervals.
Exploration must also be adapted to the queue state, since repeated exploration in the presence of new arrivals can cause backlog to accumulate and persist into subsequent rounds.

We address these difficulties by a parameter-free, queue-dependent policy, which collects observations through uniform exploration when the queue is empty, exploits the resulting estimates when a new busy period begins.
When a queue-dependent test indicates that continued greedy exploitation may be unsafe, it switches to a robust bandit policy, Exp3-IX~\citep{neu2015explore}.

\paragraph{Contributions.} 
In the oblivious adversarial regime with non-negative arrivals and bounded services, our policy achieves $\tilde{\gO}(\sqrt{K L_T})$ worst-case queue length regret with high-probability for the length of the longest busy period $L_T$.
Since $L_T\leq T$, this matches the $\Omega(\sqrt{KT})$ lower bound up to logarithmic factors, while providing a sharper guarantee when busy periods are short.
Under i.i.d.~bounded arrivals and services with a unique stabilizing best channel, the same policy obtains $\gO(1)$ expected cumulative regret and also $\gO(\log(KT)/\Delta_{\min})$ expected worst-case regret.
Hence, combined with the adversarial result, it establishes a best-of-both-worlds guarantee.
Table~\ref{tab: regret} summarizes our results and those of prior work.

\section{Preliminaries}\label{sec: prelim}
We consider a single queue served by $K\geq 2$ channels over rounds $t\in [T] = \{1,\ldots, T\}$. 
Let $Q_t^\pi \geq 0$ denote the queue length at the beginning of round $t$ under a policy $\pi$, with $Q_1^\pi=0$.
At each round $t$, an amount $A_t \geq 0$ of data arrives, and channel $i\in[K]$ offers a service $R_{t,i}\in [0,1]$.
In this paper, we adopt an arrivals-first convention, where $A_t$ enters the queue before the service~\citep{borodin2001adversarial, kim2024queueing}.
We additionally assume that $A_t$ is revealed before the scheduling decision.
At each round $t$, based on $Q_t^\pi, A_t$ and previous observations, the policy $\pi$ selects a channel $i_t\in[K]$ without knowing $R_t$, and observes only the service $R_{t,i_t}$ of the selected channel.
Then, the queue length is updated according to
\begin{equation}\label{def: queue lindley}
  Q_{t+1}^\pi = (Q_t^\pi+ A_t - R_{t,i_t})^+,
\end{equation}
where $(x)^+:=\max(x,0)$.
Following the previous work \citep{stahlbuhk2021learning,freund2024efficient}, we assume $R_{t,i_t}$ is observable even when $Q_t^\pi+A_t=0$.

\subsection{Queue length regret}
In this paper, we define an empty period as a maximal consecutive interval satisfying $Q_t^\pi+A_t=0$, which is more restrictive than previous approaches based only on $Q_t^\pi=0$.
A busy period begins at round $\beta$ with $Q_\beta^\pi=0$ and $A_\beta>0$, and consists of the rounds from $\beta$ up to, but not including, the next round $t>\beta$ at which $Q_t^\pi=0$.
If no such round exists, the busy period continues until $T$.
We denote the busy periods chronologically by $(\gB_p)_{p\geq 1}$ and the empty period immediately preceding $\gB_p$ by $\gE_p$.
Note that consecutive busy periods $\gB_p, \gB_{p+1}$ may be adjacent if the queue is cleared in one round and a positive arrival occurs in the next.
In this case, they are separated only by the empty boundary state $Q_t^\pi =0$, and we set the intervening empty period $\gE_{p+1}=\emptyset$ for notational consistency.
Thus, the sequence of periods is represented by $(\gE_1,\gB_1,\gE_2,\ldots)$.
We define the longest length of busy period up to $T$ by
\begin{equation}\label{def: maximum length}
  L_T := \max\qty{|\gB_p\cap [T]|: p \in \sN}.
\end{equation}
For $i\in[K]$, let $Q_t^i$ be the queue length under the constant policy that always selects $i$, so that $Q_{t+1}^i = (Q_t^i + A_t - R_{t,i})^+$.
The worst-case queue length regret is defined by
\begin{equation}\label{def: worst-case regret}
  \Reg_\pi^{\max}(T) := \max_{t\in[T]} \qty(\max_{i\in[K]} Q_t^\pi - Q_t^i).
\end{equation}
In the adversarial regime, we assume the sequence $((A_t, R_t))_t$ is determined by an oblivious adversary in advance satisfying $A_t\geq0$ and $R_t\in[0,1]^K$.
The following lemma shows that the queue length regret is bounded by the cumulative service regret within the current busy period rather than over the entire horizon.
\begin{lemma}[Lemma~1 of \citet{krishnakumar2025minimizing}]\label{lem: queue to service regret}
    For any $p\in \sN$, let $\beta_p:= \min \gB_p$ denote the first round of $\gB_p$. Then, for any $t\in \gB_p$
    \begin{equation*}
        Q_t^\pi - \min_{i\in [K]}Q_t^i \leq \max_{i\in [K]} \sum_{\tau=\beta_p}^{t-1} (R_{\tau,i}-R_{\tau,i_\tau}).
    \end{equation*}
\end{lemma}
This result also explains why no upper bound on $A_t$ is imposed in the adversarial regret.
Although large $A_t$ can produce large queue lengths, these terms cancel when queue lengths are compared since the policy and every constant policy receive the same arrivals.

In the stochastic regime, we assume the sequence $((A_t, R_t))_t$ is i.i.d., where $A_t\in[0,1]$ and $R_t \in[0,1]^K$, and $A_t$ is independent of $R_t$.
We allow arbitrary bounded distributions and fractional arrivals and services, which encompasses the Bernoulli model in \citet{stahlbuhk2021learning}.
Instead, we only require 
\begin{equation}\label{asm: lb on rho}
  \Pr(A_t=0)=\rho >0, \,\forall t \in \sN,
\end{equation}
where $\rho$ is a problem-dependent constant unknown to the policy.
We denote their expectations~by
\begin{equation*}
  \lambda := \E[A_t], \quad \mu_i := \E[R_{t,i}], \quad \forall i \in [K], \forall t \in \sN.
\end{equation*}
We assume that the best channel $i^*:=\argmax_{i\in[K]}\mu_i$ is unique and stabilizing, i.e., $\epsilon:=\mu_{i^*}-\lambda>0$.
We define suboptimality gaps by $\Delta_{i}= \mu_{i^*} - \mu_i$ with $\Delta_{\min} = \min_{i\ne i^*} \Delta_{i}$.
The expected cumulative queue length regret is defined as
\begin{equation}\label{def: cumulative queue regret}
  \Reg^{\mathrm{cum}}_\pi (T) := \sum_{t=1}^{T} \E\qty[Q_t^\pi - Q_t^{i^*}].
\end{equation}
The conditions $\rho>0$ and $\epsilon >0$ are needed to guarantee bounded cumulative queue length regret in general.
If $\rho=0$, a multi-armed bandit instance gives an $\Omega(\log T)$ lower bound.
Without $\epsilon>0$, every policy can incur $\Omega(T)$ cumulative regret~\citep{stahlbuhk2021learning}.

\subsection{Previous approaches}
Since each channel can be viewed as a bandit arm with loss $\ell_{t,i} = 1- R_{t,i}$, a natural baseline is to apply a bandit policy directly to the observed services.
However, even in the stochastic regime, such queue-independent policies can incur $\Omega(\log T)$ cumulative queue length regret~\citep{stahlbuhk2021learning}.
To exploit the queueing structure, \citet{stahlbuhk2021learning} construct a queue-dependent three-phase policy whose cumulative regret is bounded uniformly over $T$ and thus converges to a finite, problem-dependent constant under Bernoulli models whenever at least one channel is stabilizing.

In particular, at the first round of each empty period, they sample an arm uniformly at random and update its empirical  estimate, after which it remains idle until the queue becomes busy.
During the $p$-th busy period, they select the empirical best arm, based on observations during empty periods, at most $p$ times.
If the queue remains busy even after these rounds, the policy enters a recovery phase that alternates between uniform-exploration rules and greedy selection based on the local estimates until the queue empties.

In contrast, \citet{krishnakumar2025minimizing} study the worst-case queue length regret under arrivals and services selected by an oblivious adversary. 
They reduce the queueing problem to controlling adversarial bandit regret uniformly over all subintervals and introduce a queue-independent policy that combines online gradient descent with importance-weighted (IW) loss estimates and uniform exploration, using knowledge of $T$ to tune its exploration rates.
It achieves a high-probability regret bound of $\tilde{\gO}(\sqrt{K}T^{3/4})$, leaving a gap to the $\Omega(\sqrt{KT})$ lower bound.

Combining these guarantees requires controlling the backlog accumulated due to greedy exploitation in adversarial regimes while ensuring that recovery occurs sufficiently rare in stochastic regimes.
The time-out rule of \citet{stahlbuhk2021learning} allows increasingly long exploitation rounds as the busy period index grows, which may lead to large regret in adversarial regimes.
Lemma~\ref{lem: queue to service regret} suggests that the queue state can help address these difficulties by restricting the relevant intervals to individual busy periods.

\section{Proposed Policy}\label{sec: policy}
We propose a parameter-free, queue-dependent policy with three phases, following the structure of \citet{stahlbuhk2021learning}.
The policy requires only the number of arms $K$ without knowledge of any environmental parameters or the horizon $T$.
Its key components are an additional queue-dependent condition for exploitation and an adversarial bandit policy for recovery.
During empty periods, our policy explores arms uniformly and updates its empirical loss estimates.
Then, each busy period begins  by exploiting the empirical best arm identified from these observations.
If the prescribed conditions fail before the queue empties, it enters the recovery phase by initializing an Exp3-IX instance~\citep{neu2015explore}, which continues until the queue becomes empty.

The queue-dependent condition is introduced to control the excess backlog before recovery, while Exp3-IX controls the additional service regret during recovery.
Initializing a fresh instance in each recovery phase allows this regret to be analyzed over its local rounds within the current busy period.

\subsection{Exploration phase}
Whenever $Q_t^\pi+A_t=0$, the policy enters the exploration phase, where we count all exploration rounds by $m\in \sN$.
At the $m$-th exploration round, corresponding to global round $t$, the policy draws $i_t$ uniformly from $[K]$ and observes $R_{t,i_t}$. 
It then constructs an IW loss estimator by $\hat{\ell}_{m,i}^{\emptyset} = K\I[i_t=i](1-R_{t,i_t})$ and updates the cumulative loss estimate by $\hat{L}_{m+1}^{\emptyset} = \sum_{\tau=1}^m \hat{\ell}_{\tau}^{\emptyset}$ with $\hat{L}_{1}^{\emptyset}=\mathbf{0}$.
These estimates are updated only during exploration phase and subsequently used to identify the empirical best arm at the beginning of each busy period.

\subsection{Exploitation phase}
At the beginning of $\gB_p$, where $Q_t^\pi=0$ and $A_t>0$, the policy enters the exploitation phase and initializes the local exploitation round to $n=1$.
In this phase, the policy always selects
\begin{equation}\label{def: empirical best}
  \hat{i}^*_p = \argmin_{i\in [K]} \hat{L}_{m+1,i}^{\emptyset}, \text{ where } m = \sum_{\tau=1}^p |\gE_\tau|.
\end{equation}
To determine whether this greedy selection is safe, we introduce an auxiliary optimistic queue length $O_t$ with $O_1=0$ updated in every round according to
\begin{equation*}
  O_{t+1} = (O_t + A_t - 1)^+.
\end{equation*}
Since $R_{t,i} \leq 1$, $O_t\leq Q_t^\pi$ holds.
Hence, $Q_t^\pi-O_t$ measures the excess backlog incurred by imperfect service. 

At the $n$-th exploitation round of $\gB_p$, corresponding to global round $t$, the policy keeps selecting $\hat{i}^*_p$ if
\begin{equation}\label{def: threshold}
  \I[n\leq p, Q_t^\pi-O_t\leq 4\sqrt{n\log(p+1)} =: g_p(n)]=1.
\end{equation}
The first condition is the time-out rule of \citet{stahlbuhk2021learning}, and the second controls the excess backlog, which is essential for our adversarial guarantee.
Note that both conditions hold automatically at $n=1$, since $p\geq1$ and $Q_{\beta_p}=O_{\beta_p}=0$.
If the queue becomes empty, the exploitation phase ends, and if either condition fails while the queue remains busy, the policy transitions to the recovery phase.

\begin{remark}[Observation of arrivals]
Under our arrivals-first convention in discrete-time scheduling, $Q_t^\pi$ represents the backlog carried into round $t$ before new work enters, while $A_t$ represents the work accumulated since the beginning of round $t-1$. 
At the beginning of round $t$, this work is admitted before the scheduling decision and service.
We assume that the scheduler can observe whether the resulting queue is empty.
In other words, our policy requires only the indicator $\I[Q_t^\pi+A_t=0]$ and this is a queue-state observation at the beginning of round $t$ and \emph{does not provide any future information} about arrivals or services.

This indicator distinguishes an exploration opportunity, where $Q_t^\pi=A_t=0$, from the beginning of a busy period, where $Q_t^\pi=0$ and $A_t>0$.
Hence, the policy can use genuinely empty rounds to improve its estimations in preparation for the next busy period and then select the empirical best arm $\hat{i}_p^*$ from its first round $\beta_p$.
In particular, the indicator allows the policy to respond to newly admitted work without a one-round delay.
If the decision were based only on $Q_t^\pi$, the policy could not distinguish these two cases and might perform uniform exploration at round $\beta_p$, selecting the empirical best arm only in the following round. 
Such unnecessary exploration can create additional queue length regret even when the optimal arm has already been identified correctly.
This immediate transition from exploration to exploitation is needed for our stochastic cumulative regret guarantee. 
Further details are given in Appendix~\ref{app: why arrival}.

Indeed, the need for this assumption in our policy depends on the queue dynamics.
Our definition in (\ref{def: queue lindley}) follows the arrivals-first convention adopted in much of the queueing bandit literature most closely related to our setting~\citep{krishnasamy2016regret,krishnasamy2021learning,kim2024queueing,freund2024efficient, krishnakumar2025minimizing,bae2026queue}.
In comparison, \citet{stahlbuhk2021learning} study cumulative queue length regret under the service-first convention
\begin{equation}\label{def: queue stan}
  Q_{t+1}^\pi=(Q_t^\pi-R_{t,i_t})^+ + A_t,
\end{equation}
which is used in the stochastic network optimization literature~\citep{neely2010stochastic}.
In their model, $Q_t^\pi=0$ implies $Q_{t+1}^\pi=A_t$, regardless of the selected arm.
Therefore, exploration during empty periods preserves the queue length difference without observing the current arrival before decision.
Our arrival-observation assumption serves an analogous purpose under the dynamics in (\ref{def: queue lindley}).
\end{remark}

\subsection{Recovery phase}
Once the policy enters the recovery phase, then it remains in this phase for the rest of the current busy period.
At the beginning of the recovery phase in $\gB_p$, it initializes an Exp3-IX instance with local round counter $s=1$ and cumulative estimated losses $\hat{L}_1^{\mathrm{ix}}=\mathbf{0}$.
At the $s$-th recovery round, corresponding to global round $t$, we define for all $i\in [K]$
\begin{equation}\label{def: eta and zeta}
    \eta_s = \sqrt{\frac{\log K}{Ks}}, \text{ and } w_{s,i} = \frac{\exp(-\eta_s \hat{L}_{s,i}^{\mathrm{ix}})}{\sum_{j=1}^K \exp(-\eta_s \hat{L}_{s,j}^{\mathrm{ix}})}.
\end{equation}
The policy samples $i_t \sim w_s$, observes $R_{t,i_t}$, constructs the implicit-exploration loss estimator
\begin{equation}\label{def: ix loss}
  \hat{\ell}_{s,i}^{\mathrm{ix}} = \frac{\I[i_t=i](1-R_{t,i_t})}{w_{s,i}+\zeta_s},
\end{equation}
and then updates $\hat{L}_{s+1}^{\mathrm{ix}}=\sum_{\tau=1}^s \hat{\ell}_{\tau}^{\mathrm{ix}}$.
Here, $\zeta_s = \eta_s/2$ is the implicit-exploration parameter introduced by \citet{neu2015explore}.
Adding $\zeta_s$ introduces a controlled bias and limits the magnitude of the loss estimator for small $w_{s,i}$, enabling high-probability adversarial regret guarantees without explicit uniform exploration.

The Exp3-IX instance runs until the queue becomes empty, after which its local estimates are discarded.
Thus, recovery observations are not shared across busy periods and not incorporated into the exploration estimate $\hat{L}_m^{\emptyset}$. 
While \citet{stahlbuhk2021learning} employed UCB-type policies in their heuristic policy, our use of Exp3-IX is designed to provide guarantees under both stochastic and adversarial regimes.
Specifically, its high-probability regret bound controls the additional service regret during recovery in the adversarial regime.
The complete pseudo-code is given in Algorithm~\ref{alg: policy}.

\begin{algorithm*}[!t]
\small
\caption{Best-of-Both-Worlds Bandit Scheduling}
\label{alg: policy}
\DontPrintSemicolon
\textbf{Input:} $K\geq2$; Initialize $t=1$, $Q_1 = O_1 = 0$, $m=p=0$, $\hat{L}_1^{\emptyset} = \mathbf{0}$.

\While{True}{

Observe the arrival $A_t$.

  \If(\tcp*[f]{Empty period: Exploration phase}){$Q_t+A_t=0$}{
      Update exploration round $m \gets m+1$.

      Play $i_t \sim \mathrm{Uniform}(K)$ and observe service $R_{t,i_t}$.

      Set $\hat{\ell}_{m,i}^{\emptyset}=\I[i_t = i] K(1-R_{t,i_t})$ and update $\hat{L}_{m+1}^{\emptyset} = \hat{L}_{m}^{\emptyset} + \hat{\ell}_m^{\emptyset}$.
  }
  \Else(\tcp*[f]{Busy period}){
      \If(\tcp*[f]{First round of Exploitation phase}){$Q_t=0$}{
            
      Set $\mathsf{phase} \gets \mathsf{exploitation}$.

      Increase the busy period count $p \gets p+1$, reset exploitation round $n\gets 1$. 

      Take the empirical best $\hat{i}^*_p = \argmin_{i \in [K]} \hat{L}_{m+1,i}^{\emptyset}$ in empty periods.

      Play $i_t = \hat{i}^*_p$ and observe service $R_{t,i_t}$.

      Update exploitation round $n \gets 2$.
      }

      \Else(\tcp*[f]{When $Q_t^\pi > 0$}){

      \If(\tcp*[f]{Exploitation phase})%
      {$\mathsf{phase}=\mathsf{exploitation}$}{
        \If{$n \leq p$, $Q_{t}-O_{t}\leq g_p(n)$}{
          Play $i_t = \hat{i}^*_p$, observe service $R_{t,i_t}$, and update exploitation round $n \gets n+1$.}
        \Else(\tcp*[f]{Initialize the recovery phase}){
          Set $\mathsf{phase} \gets \mathsf{recovery}$.
        
          Initialize a new Exp3-IX instance by $s\gets 1$ and $\hat{L}_{1}^{\mathrm{ix}}=\mathbf{0}$.
        }
      }
      \If(\tcp*[f]{Recovery phase: Exp3-IX}){$\mathsf{phase}=\mathsf{recovery}$}{
        
        Set $\eta_s$ and $w_s$ as in (\ref{def: eta and zeta}), and $\zeta_s = \eta_s/2$.

        Play $i_t \sim w_s$ and observe service $R_{t,i_t}$.

        Set $\hat{\ell}_{s,i}^{\mathrm{ix}}$ as in (\ref{def: ix loss}) and update $\hat{L}_{s+1}^{\mathrm{ix}} = \hat{L}_{s}^{\mathrm{ix}} + \hat{\ell}_{s}^{\mathrm{ix}}$ and recovery round $s \gets s+1$.
        }
       } }

    Update queue length $Q_{t+1} = (Q_{t}+A_t - R_{t,i_t})^+$ and $O_{t+1} = (O_{t}+A_t - 1)^+$.
    
    Update global round $t \gets t+1$.
}
\end{algorithm*}

\section{Theoretical Guarantee}\label{sec: result}
In this section, we provide the regret bounds of our policy in both adversarial and stochastic regimes, whose detailed proofs are given in Appendices~\ref{app: adv},~\ref{app: sto}, and~\ref{app: cor}.
\begin{theorem}\label{thm: adv}
  In the oblivious adversarial regime, for any $T\geq 1$ and $\delta \in (0,1)$, Algorithm~\ref{alg: policy} satisfies 
  \begin{equation*}
    \Reg_{\pi}^{\max}(T) \leq 4\sqrt{L_T\log (T+1)} + 4\sqrt{KL_T \log K} + \qty(2\sqrt{\frac{KL_T}{\log K}}+1) \log(\frac{6L_T^2T^2}{\delta}) +3
  \end{equation*}
  with probability at least $1-\delta$ whenever $L_T\geq 1$.
\end{theorem}
Note that $L_T=0$ means that the queue remains empty for all rounds, where the worst-case queue length regret is trivially zero. Moreover, since $\Reg_{\pi}^{\max}(T)\leq T$ holds by $R_{t} \in [0,1]^K$ and Lemma~\ref{lem: queue to service regret}, setting $\delta=1/T$ gives an expected regret bound of the same order, up to logarithmic factors.

\begin{proof}[Proof sketch]
Since threshold $g_p(n)$ controls the excess backlog accumulated during exploitation and Exp3-IX controls the additional regret incurred during recovery, the analysis reduces to controlling the regret of each recovery instance at every local round.

Let $\Reg_{\mathrm{IX}}(s;p)$ be the regret of corresponding Exp3-IX instance accumulated over the first $s$ recovery rounds during $p$-th busy period.
Then, we establish that
\begin{equation*}
  \Reg_{\pi}^{\max}(T) \lesssim \sqrt{L_T \log T} + \max_{p, s} (\Reg_{\mathrm{IX}}(s;p))^+.
\end{equation*}
The first term follows the exploitation threshold in (\ref{def: threshold}), which controls the worst-case regret when the policy enters recovery.
For the second term, \citet{neu2015explore} provides a high-probability regret bound of $\tilde{\gO}(\sqrt{KH})$ for an Exp3-IX instance over $H$ rounds.
We apply this bound to each instance and take a union bound over all instance indices $p$ and local rounds $s$.
Since every recovery instance lasts at most $L_T$ rounds, we obtain the desired high-probability regret bound.
\end{proof}
In the stochastic regime, the cumulative regret is finite if the unique best arm is stabilizing and both the arrivals and services are i.i.d.~with support $[0,1]$.
\begin{theorem}\label{thm: sto cum}
    Suppose that $((A_t,R_t))_{t}$ is i.i.d.~with $A_t\in[0,1]$, $\Pr(A_t=0)=\rho>0$, and $R_t\in[0,1]^K$, and that the unique best arm $i^*$ is stabilizing with $\epsilon = \mu_{i^*}-\lambda>0$. Then, Algorithm~\ref{alg: policy} satisfies
    \begin{equation*}
        \sum_{t=1}^\infty \E\qty[\qty(Q_t^\pi - Q_t^{i^*})^+] \leq C_{\mathrm{sto}} <\infty
    \end{equation*}
    where 
    \begin{equation*}
      C_{\mathrm{sto}} = \gO\qty(\frac{K^4}{\epsilon^2 \Delta_{\min}^6 \rho^3} + \frac{K^4 \log^2 K}{\epsilon^4 \Delta_{\min}^2 \rho} + \frac{1}{\epsilon^{8}}+ \frac{K^2 \log^2 K}{\epsilon^6}).
    \end{equation*} 
\end{theorem}

\begin{proof}[Proof sketch]
For the cumulative queue length regret, we follow the busy period decomposition of \citet{stahlbuhk2021learning}.
We first define the good period indicator
\begin{equation*}
  \gJ_p :=\I\qty[\hat{i}_p^* =i^*, \gB_p^{\mathrm{ix}}=\emptyset ],
\end{equation*}
where $\gB_p^{\mathrm{ix}}$ denotes the interval of recovery phase during the busy period $\gB_p$.
When $\gJ_p=1$, the policy selects $i^*$ throughout $\gB_p$, including its first round $\beta_p$.
Since the empty period is determined by the queue length of the policy, $Q_{\beta_p}^{i^*}$ can be positive and the monotonicity of the queue recursion implies $Q_t^\pi \leq Q_t^{i^*}$ throughout this period.
Empty periods also contribute no positive queue length difference.
Hence, it suffices to control the contributions of busy periods satisfying $\gJ_p=0$.

The argument combines a bound on the regret in $\gJ_p=0$ with a bound on its probability.
Since bounded arrivals imply $O_t=0$, $g_p(n)$ directly controls the queue length, while the time-out condition limits greedy exploitation to at most $p$ rounds.
During the recovery phase, the Exp3-IX regret bound and the stability gap $\epsilon$ provide tail bounds on the rounds required to empty the queue.
By combining these arguments, we obtain
\begin{equation*}
  \E\qty[ \sum_{t\in\gB_p}(Q_t^\pi-Q_t^{i^*})^+] \leq C_p\Pr(\gJ_p=0),
\end{equation*}
where $C_p$ is polynomial in $p$, $\epsilon^{-1}$, and $K$.

By definition, $\gJ_p=0$ occurs only if $\hat{i}^*_p$ is incorrect or exploitation phase ends before the queue becomes empty.
Since the assumption $\Pr(A_t=0)=\rho>0$ provides certain degree of exploration opportunities, combining a concentration inequality on the number of exploration rounds, with a time-uniform version of Freedman's inequality for the loss estimates, controls the probability of selecting an incorrect empirical best arm.
When the empirical best arm is correct, Hoeffding's inequality and $\epsilon>0$ controls the probabilities of violating the queue length threshold or reaching the time-out before the queue empties.
The obtained probability bound decays sufficiently fast so that
\begin{equation*}
  \sum_{p=1}^{\infty}C_p\Pr(\gJ_p=0) \leq C_{\mathrm{sto}}<\infty,
\end{equation*}
which concludes the proof.
\end{proof}
However, Theorem~\ref{thm: sto cum} does not directly bound the expected worst-case queue length regret, since the maximum over time is taken before the expectation. 
The following corollary provides such a guarantee.
\begin{corollary}\label{cor}
Under the assumptions of Theorem~\ref{thm: sto cum}, for any $T\geq1$, Algorithm~\ref{alg: policy} satisfies
  \begin{equation*}
    \E[\Reg_{\pi}^{\max}(T)] \leq C_{\mathrm{sto}} + \gO\qty(\frac{\log (KT)}{\Delta_{\min}}).
  \end{equation*}
\end{corollary}
Thus, our policy simultaneously achieves uniformly bounded expected cumulative regret and logarithmic expected worst-case regret in the stochastic regime.
\begin{proof}[Proof Sketch]
For worst-case regret, it suffices to additionally compare $Q_t^{i^*}$ with every constant policy as
\begin{equation*}
  \Reg_\pi^{\max}(T) \leq \max_{t}(Q_t^\pi-Q_t^{i^*})^+ + \max_{t, i} (Q_t^{i^*}-Q_t^i)^+.
\end{equation*}
The expected first term is bounded by $C_{\mathrm{sto}}$.
For each suboptimal arm $i$, the service difference $R_{t,i}-R_{t,i^*}$ has negative mean $-\Delta_i$.
Applying concentration inequalities to these service differences over all relevant intervals provides the bound on the second term,
\begin{equation*}
  \E[\Reg_\pi^{\max}(T)] \leq C_{\mathrm{sto}} +\gO\left(\frac{\log(KT)}{\Delta_{\min}}\right),
\end{equation*}
which concludes the proof.
\end{proof}

\section{Numerical experiments}\label{sec: exp}
To verify our best-of-both-worlds guarantee, we evaluate our policy in an adversarially varying environment and an i.i.d.~stochastic environment, conducting $1000$ independent runs of $T=10000$ rounds in each setting.
In all figures, the lines denote the empirical means and shaded regions indicate pointwise $95$\% confidence intervals.
Further experimental details are in Appendix~\ref{app: exp}.

\textbf{Baselines.}
We compare our policy, Algorithm~\ref{alg: policy}, with (i) Q-ThS~\citep{krishnasamy2016regret}, which is designed to minimize the queue length regret, (ii) the three phase policy of \citet{stahlbuhk2021learning}, which is designed for cumulative queue length regret, and (iii) OGD-based policy of \citet{krishnakumar2025minimizing}, which is designed to minimize worst-case queue length regret.
In this section, we refer to the latter two policies as SSM-3P and KS-OGD, respectively.

\textbf{Adversarial environments.}
In this setting, we follow the instance considered in \citet{krishnakumar2025minimizing}, where there are $K=5$ non-stationary arms and we divide the horizon into $h=7$ consecutive blocks of equal lengths.
At the beginning of each block, we sample a coefficient $\mu_{i}^{(h)}$ from $\mathrm{Uniform}(0,1)$ and fix this value.
Then, the services of $i$ at every $t+1$ in block $h$ is determined by
\begin{equation*}
  R_{t+1,i} = \max(0, \min(1, \mu_i^{(h)}\cdot R_{t,i} + \xi_{t,i} )),
\end{equation*}
where $\xi_{t,i} \sim \mathrm{Uniform}(-1,1)$ is an i.i.d.~additive noise.
The arrivals are sampled from a uniform distribution with mean $\lambda$ satisfying $\mu_{i_T^*} - \lambda= 0.05$. 

\begin{figure*}[t]
    \centering
    \begin{subfigure}[t]{0.43\textwidth}
        \centering
        \includegraphics[width=\linewidth]
        {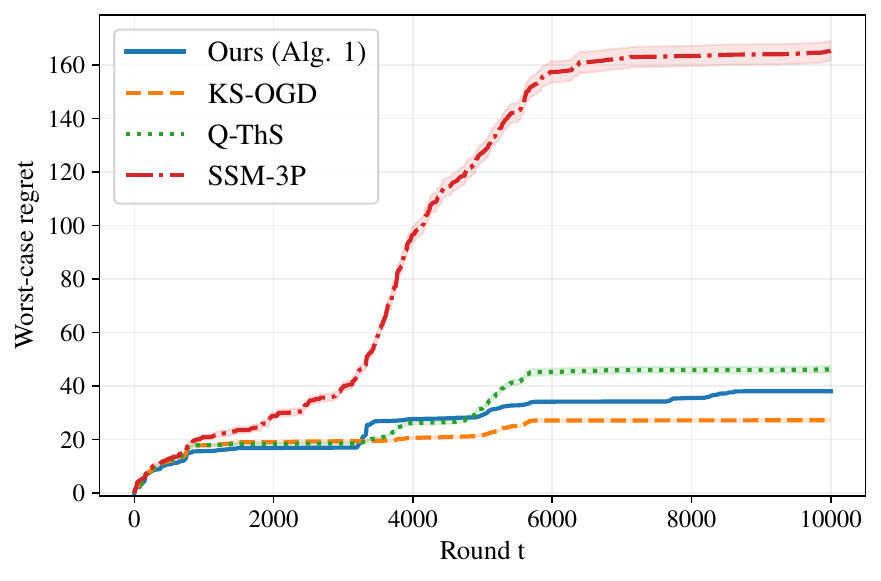}
        \caption{Worst-case regret.}
        \label{fig:adv worst}
    \end{subfigure}
    \quad
    \begin{subfigure}[t]{0.43\textwidth}
        \centering
        \includegraphics[width=\linewidth]
        {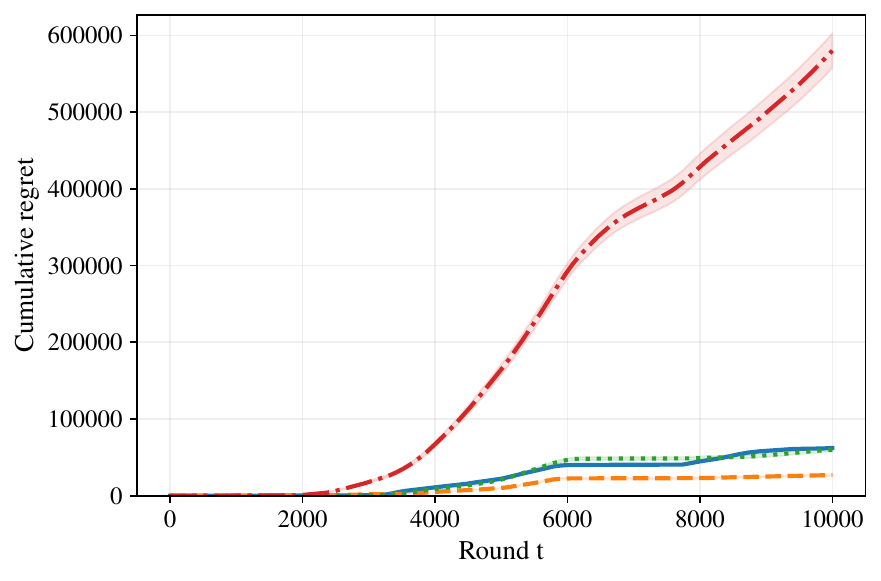}
        \caption{Cumulative regret.}
        \label{fig:adv cum}
    \end{subfigure}
    \caption{Results in the adversarial regime.} 
    \label{fig: adv}
\end{figure*}

Figure~\ref{fig: adv} shows the empirical queue length regret in this adversarial instance.
KS-OGD performs best on this instance, whereas SSM-3P performs worst, consistent with the observations in \citet{krishnakumar2025minimizing}.
Our policy achieves performance comparable to KS-OGD and Q-ThS in both worst-case and cumulative queue length regret.

\begin{figure*}[t]
    \centering
    \begin{subfigure}[t]{0.43\textwidth}
        \centering
        \includegraphics[width=\linewidth]
        {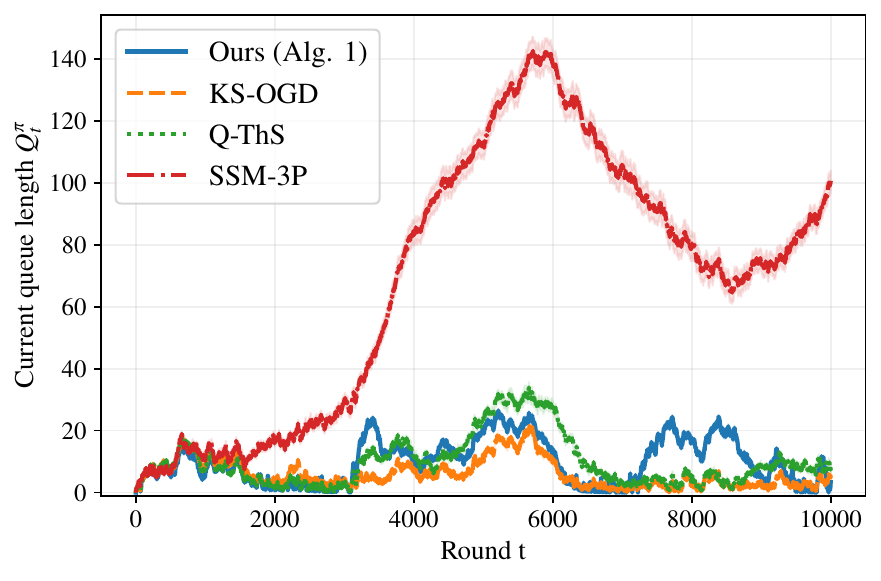}
        \caption{Adversarial regime.}
        \label{fig:adq}
    \end{subfigure}
    \quad
    \begin{subfigure}[t]{0.43\textwidth}
        \centering
        \includegraphics[width=\linewidth]
        {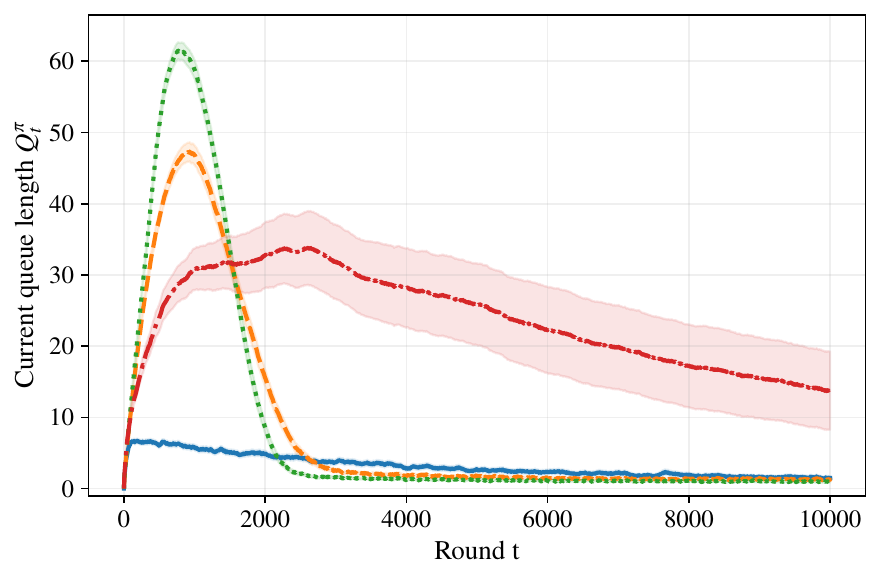}
        \caption{Stochastic regime.}
        \label{fig:stq}
    \end{subfigure}
    \caption{The trajectories of queue lengths $Q_t^\pi$.}
    \label{fig: queue}
\end{figure*}

The queue length trajectories in Figure~\ref{fig:adq} explicitly show the increase after some block transitions, particularly between rounds $4000$ and $6000$, i.e., the fourth block.
All four policies experience substantial regret increase during this interval, suggesting that these transitions introduce significant environmental changes.
For our policy, such changes may make the arm selected for greedy exploitation suboptimal, resulting in additional backlog before the recovery phase.

\begin{figure*}[t]
    \centering
    \begin{subfigure}[t]{0.43\textwidth}
        \centering
        \includegraphics[width=\linewidth]
        {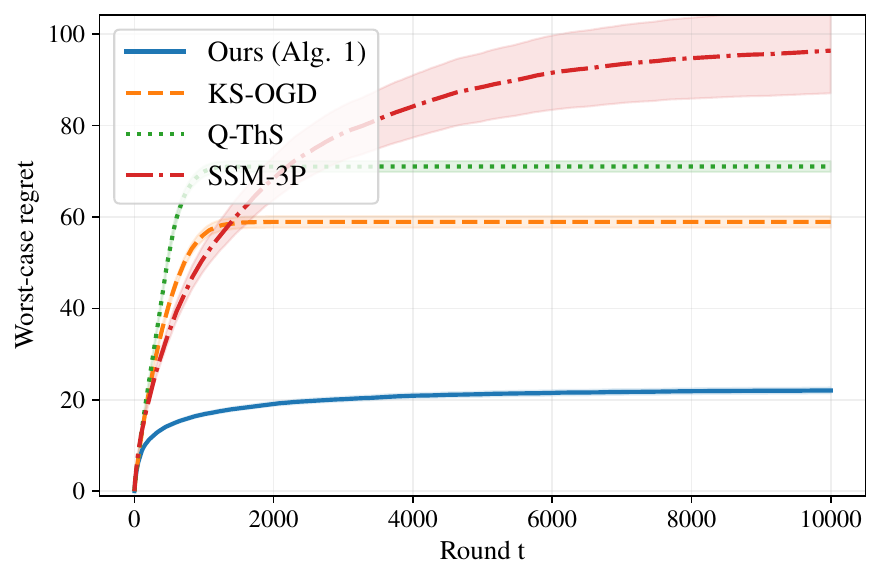}
        \caption{Worst-case regret.}
        \label{fig:st worst}
    \end{subfigure}
    \quad
    \begin{subfigure}[t]{0.43\textwidth}
        \centering
        \includegraphics[width=\linewidth]
        {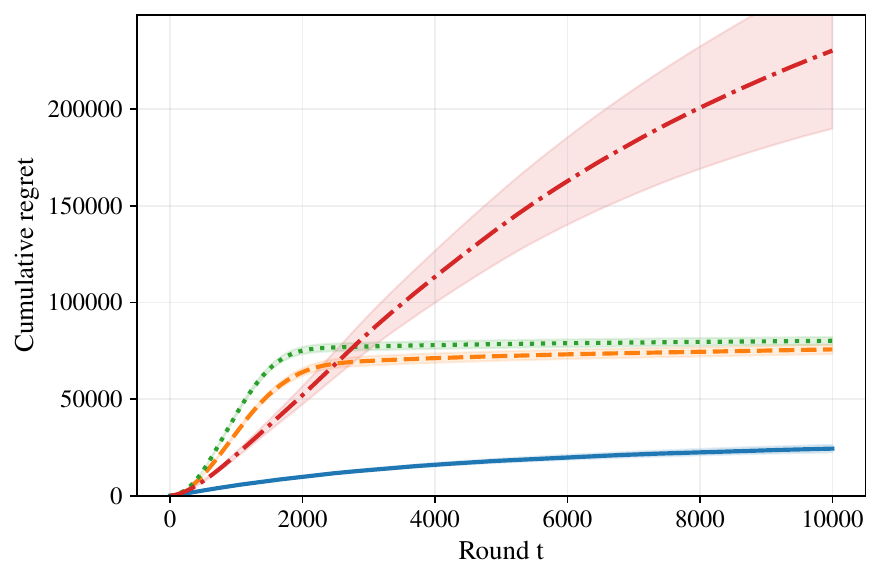}
        \caption{Cumulative regret.}
        \label{fig:st cum}
    \end{subfigure}
    \caption{Results in the stochastic regime.} 
    \label{fig: st}
\end{figure*}

\textbf{Stochastic environments.}
For the stochastic regime, we follow the setting of \citet{stahlbuhk2021learning}, with i.i.d.~Bernoulli arrivals and services of expectations $\lambda=0.5$ and $\mu=(0.1,0.3,0.5,0.7)$, respectively.
As shown in Figure~\ref{fig: st}, our policy achieves substantially lower regret than the baselines under both regrets.
Moreover, our policy, KS-OGD, and Q-ThS appear to converge to finite values as the horizon increases.
Combined with the adversarial experiments, these results complement our theoretical analysis by illustrating the competitive performance of our policy across both regimes.

The queue length trajectories in Figure~\ref{fig:stq} provide further insight into these differences.
In particular, Q-ThS and KS-OGD exhibit a clear transition from initial queue growth to subsequent stabilization, qualitatively resembling the two-stage behavior of queue length regret studied by \citet{krishnasamy2016regret}.
During the early stage, exploration and inaccurate service estimates can lead to an increase the queue length.
As the policy learns to select the best arm more often, it can stabilize the queue and thus queue length decreases.

In contrast, two phase-based policies, ours and SSM-3P, exhibit a less distinct separation between the two stages and a gradual decrease in later rounds.
While our policy initially has the shortest queue and reduces it first, $Q_t^\pi$ of KS-OGD and Q-ThS decrease rapidly and eventually fall below that of ours.
The gradual decrease under our policy may result from additional explorations needed for robustness, which can select suboptimal arms and delay queue clearance in stochastic regimes.
\section{Conclusion}\vspace{-0.1em}
We proposed a parameter-free policy that achieves best-of-both-worlds guarantees for single queueing bandits.
Our policy achieves $\widetilde{\gO}(\sqrt{K L_T})$ worst-case queue length regret with high-probability in the oblivious adversarial regime, where $L_T$ denotes the length of the longest busy period.
Under i.i.d.~bounded sequences with a unique stabilizing optimal arm, it simultaneously achieves constant expected cumulative queue length regret and logarithmic expected worst-case queue length regret.
These results show that a simple policy can retain strong stochastic guarantees as well as adversarial guarantees, without knowledge of the regime.
Numerical results support our analysis by showing competitive performance across both regimes.

A promising direction is to investigate whether ideas from decoupled bandits can improve the use of exploration opportunities~\citep{rouyer2020decoupled, kim2026follow}.
Our policy already separates selections for exploration from those for exploitation, assigning these roles to empty and busy periods, respectively.
This structure suggests replacing uniform exploration with adaptive sampling that accounts for uncertainty in the estimates and their importance to subsequent decisions.
Such an approach may reduce the cost of initial learning and improve practical performance.
Establishing whether such improvements can be achieved while preserving both the stochastic guarantee and adversarial robustness is an interesting direction for future work.

\subsection*{AI use statement}
In this work, we used generative AI tools, Chat-GPT 5.6 Sol xHigh, to help develop the conceptual framework through a survey of the queueing bandit literature, formulate mathematical claims, particularly notation and definitions, suggest ingredients for proofs, particularly for understanding and adapting the formulation of \citet{howard2020time} to our one-dimensional notation, assist in writing proofs through choices of terminology and notation, refine research hypotheses through a survey of different queue dynamics, provide feedback on the research methodology and experimental design, implement methods in Python.
We did not use generative AI tools for interpretation of results, writing the manuscript from the beginning, translation, cleaning or reformatting datasets, generating synthetic datasets, and qualitative and thematic data analysis was not applicable to this work.

Additionally, we used the tool for brainstorming, searching for related literature, identifying relevant references, suggesting experimental settings, generating and editing software code, revising the manuscript, improving readability, and suggesting the title of the paper.

We reviewed all AI-assisted work. 
All mathematical claims were checked and written line by line by the authors. 
All AI-assisted code was reviewed and tested against the stated algorithms, and the reported experimental results were generated and verified by the authors. 
All references and literature summaries were checked against the original sources through a manual literature review. 
The authors revised and approved all AI-assisted text.
We take responsibility for the final content of this work, including text, claims, and artifacts produced with the aid of generative AI.

\subsubsection*{Acknowledgements}
JL was supported by the National Research Foundation of Korea (NRF) grant funded by the Korea government (MSIT) (No. RS-2024-00395303) and by the grant No. 2024-00460980 (IITP) funded by Korea Government (MSIT). CK was supported
by the grant Nos. 2024-00460980; and 2025-02304717 (IITP) funded by the Korea government (the Ministry of Science and ICT).

\bibliographystyle{plainnat}
\bibliography{ref}

\clearpage
\appendix
\thispagestyle{empty}

\onecolumn
\phantomsection 
\addcontentsline{toc}{section}{Appendices} 
\section*{Appendix Contents} 
\startcontents[appendix]    
\printcontents[appendix]{l}{1}{} 
\newpage

\section{Proof of Theorem~\ref{thm: adv}: adversarial regime}\label{app: adv}
For notational simplicity, we partition each busy period into two consecutive subintervals according to the phases of Algorithm~\ref{alg: policy}. 
For each busy period $\gB_p$, let $\gB_p^{\mathrm{ex}}$ denote its exploitation interval and $\gB_p^{\mathrm{ix}}$ be the subsequent recovery interval, during which an Exp3-IX instance is run.
Thus, $\gB_p = \gB_p^{\mathrm{ex}} \sqcup \gB_p^{\mathrm{ix}}$ holds, where $\gB_p^{\mathrm{ix}}$ may be empty if the queue empties during the exploitation phase.

To quantify the contribution of the recovery phase, consider any $p$ such that $\gB_p^{\mathrm{ix}}\neq\emptyset$, and let $\tau(p,s)$ denote the global round of the $s$-th recovery round in $\gB_p^{\mathrm{ix}}$.
Specifically, it satisfies
\begin{equation*}
  \tau(p,s) = \sum_{p'=1}^{p-1} |\gE_{p'}\sqcup \gB_{p'}| +|\gE_p| + |\gB_p^{\mathrm{ex}}|+ s.
\end{equation*}
For any $1\leq s\leq|\gB_p^{\mathrm{ix}}|$, define the regret of the corresponding Exp3-IX instance over its first $s$ recovery rounds by
\begin{equation}\label{def: regret ix}
  \Reg_{\mathrm{IX}}(s;p):= \max_{i\in [K]} \sum_{s'=1}^s (R_{\tau(p,s'),i} - R_{\tau(p,s'),i_{\tau(p,s')}}) = \sum_{s'=1}^s \ell_{\tau(p,s'),i_{\tau(p,s')}} - \min_{i\in [K]} \sum_{s'=1}^s \ell_{\tau(p,s'),i}.
\end{equation}
The high-probability bound for this regret is given as follows.
\begin{lemma}[Theorem 1 of \citet{neu2015explore}]\label{lem: thm 1 of neu}
  Consider any oblivious loss sequence $(\ell_t)_t$ with $\ell_t \in [0,1]^K$.
  Then, for any fixed $s\geq 1$ and any $\delta \in (0,1)$, Exp3-IX satisfies
  \begin{equation*}
    \Pr(\Reg_{\mathrm{IX}}(s; \cdot) > 4 \sqrt{Ks\log K} + \qty(2\sqrt{\frac{Ks}{\log K}}+1)\log \frac{2}{\delta}) \leq \delta.
  \end{equation*}
\end{lemma}
Since the index $p$ only identifies the busy periods, it does not appear in this per-phase bound.
We next express the worst-case queue length regret in terms of $\Reg_{\mathrm{IX}}(s;p)$, whose proof is given in Appendix~\ref{app: worst case decomposition}.
\begin{lemma}\label{lem: worst case decomposition}
  Algorithm~\ref{alg: policy} satisfies for any $T\geq 1$
  \begin{equation*}
    \Reg_{\pi}^{\max}(T) \leq 3 + 4\sqrt{L_T \log (T+1)} + \max_{p: \gB_p^{\mathrm{ix}}\cap [T]\ne \emptyset } \max_{s \leq |\gB_p^{\mathrm{ix}}\cap [T]|} (\Reg_{\mathrm{IX}}(s;p))^+.
  \end{equation*}
\end{lemma}
Note that Lemma~\ref{lem: thm 1 of neu} considers a single Exp3-IX instance at a fixed length $s$. 
Since each recovery phase initializes a fresh instance, Lemma~\ref{lem: thm 1 of neu} can be applied to $\Reg_{\mathrm{IX}}(s;p)$ for every fixed pair $(p,s)$ after conditioning on the history preceding the $p$-th recovery phase.
Hence, a union bound extends this guarantee to all Exp3-IX instances and their realized pairs as follows.
\begin{lemma}\label{lem: uniform ix bound}
For any $\delta\in(0,1)$ and for any $T\geq 1$,
  \begin{align*}
  \max_{p: \gB_p^{\mathrm{ix}}\cap [T]\ne \emptyset } \max_{s \leq |\gB_p^{\mathrm{ix}}\cap [T]|} \Reg_{\mathrm{IX}}(s;p) \leq 4\sqrt{KL_T \log K} + \qty(2 \sqrt{\frac{KL_T}{\log K}}+1) \log\qty(\frac{6L_T^2 T^2}{\delta}).
  \end{align*}
holds with probability at least $1-\delta$.
\end{lemma}
\begin{proof}[Proof of Theorem~\ref{thm: adv}]
  Combining Lemmas~\ref{lem: worst case decomposition} and~\ref{lem: uniform ix bound} concludes the proof.
\end{proof}

\section{Proof of Lemmas for Theorem~\ref{thm: adv}}\label{app: lemma for adv}

\subsection{Proof of Lemma~\ref{lem: worst case decomposition}}\label{app: worst case decomposition}
To derive the upper bounds on the worst-case queue length regret, we first introduce the relationship between queue length $Q_t^\pi$ and optimistic queue length $O_t$ defined in Section~\ref{sec: policy}.
Since
\begin{equation*}
  Q_{t+1}^\pi = (Q_t^\pi+ A_t - R_{t,i_t})^+ \qq{and} O_{t+1} = (O_t+ A_t - 1)^+,
\end{equation*}
for any $t\geq 1$, it holds that
\begin{equation}\label{eq: O and Q}
  O_t \leq Q_t^\pi, \qq{and} O_t \leq Q_t^i, \, \forall i \in [K].
\end{equation}
Define the observable excess queue length $E_t$ by
\begin{equation}\label{def: Et}
  E_t := Q_t^\pi - O_t.
\end{equation}
Then, (\ref{eq: O and Q}) and its definition imply that
\begin{equation}\label{eq: E and Q}
  Q_t^\pi - \min_{i\in [K]}Q_{t}^i \leq E_t, \qq{and} E_t \leq E_{t-1}+ 1.
\end{equation}
Then, we need following auxiliary lemmas to prove Lemma~\ref{lem: worst case decomposition}, where the proofs are postponed to Appendix~\ref{app: aux lem}.
\begin{lemma}\label{lem: Q,E,g}
  For any $t \in \gB_{p}^{\mathrm{ex}}$, whose corresponding local exploitation round is $n\geq 1$, Algorithm~\ref{alg: policy} satisfies
  \begin{equation*}
    Q_{t}^{\pi} - \min_{i\in [K]}Q_{t}^{i} \leq g_p(n), 
  \end{equation*}
  where $g_p(n) = 4\sqrt{n\log(p+1)}$ as defined in (\ref{def: threshold}).
  Moreover, if $\gB_{p}^{\mathrm{ix}} \ne \emptyset$, it holds that 
  \begin{equation*}
    E_{\tau(p,1)} \leq g_p(|\gB_p^{\mathrm{ex}}|) + 1,
  \end{equation*}
  where $\tau(p,1)$ denotes the global round of the first local recovery rounds in the $p$-th busy period.
\end{lemma}
\begin{lemma}\label{lem: Q,E,Reg}
  For any $t \in \gB_p^{\mathrm{ix}}$, whose corresponding local recovery round is $s\geq 1$, Algorithm~\ref{alg: policy} satisfies
  \begin{equation*}
    Q_t^\pi - \min_{i \in [K]} Q_t^i \leq E_{\tau(p,1)} + \Reg_{\mathrm{IX}}(s;p) +1.
  \end{equation*}
\end{lemma}
\begin{proof}[Proof of Lemma~\ref{lem: worst case decomposition}]
  Let $\max_{x \in \emptyset}f(x) = 0$ for any function $f$.
  Then, by Lemmas~\ref{lem: Q,E,g} and~\ref{lem: Q,E,Reg}, we obtain
  \begin{align*}
    \Reg^{\max}_\pi(T) &= \max_{t \in [T]}\max_{i\in [K]} (Q_{t}^\pi - Q_t^i) \\
    &= \max_{p \geq 1} \max_{\tau \in \gB_{p}\cap [T]} \max_{i \in [K]} (Q_\tau^\pi - Q_\tau^i) \tag{$\because Q_t^\pi = 0$ for $t \in \gE_{p}, \forall p\geq 1$} \\
    &= \max_{p \geq 1} \max\qty(\max_{\tau \in \gB_{p}^{\mathrm{ex}}\cap [T]} \max_{i \in [K]} (Q_\tau^\pi - Q_\tau^i), \max_{\tau \in \gB_{p}^{\mathrm{ix}}\cap [T]} \max_{i \in [K]} (Q_\tau^\pi - Q_\tau^i)) \tag{disjoint phases} \\
    &\leq \max_{p \geq 1} \max\qty(\max_{n \leq |\gB_{p}^{\mathrm{ex}}|} g_p(n), \max_{s\leq |\gB_{p}^{\mathrm{ix}}|} (E_{\tau(p,1)}+\Reg_{\mathrm{IX}}(s;p)+2)) \tag{by Lemmas~\ref{lem: Q,E,g} and~\ref{lem: Q,E,Reg}} \\
    &\leq \max_{p \geq 1} \max\qty(\max_{n \leq |\gB_{p}^{\mathrm{ex}}|} g_p(n), \max_{s\leq |\gB_{p}^{\mathrm{ix}}|} \qty(g_p(|\gB_{p}^{\mathrm{ex}}|)+3+\Reg_{\mathrm{IX}}(s;p)))  \tag{by Lemma~\ref{lem: Q,E,g}}\\
    &\leq 4\sqrt{L_T\log (T+1)} + 3 + \max_{p\geq 1} \max_{s\leq |\gB_{p}^{\mathrm{ix}}|}(\Reg_{\mathrm{IX}}(s;p))^+.
  \end{align*}
  The last inequality follows since $|\gB_{p}^{\mathrm{ex}}| \leq L_T$ by definition of $L_T$ in (\ref{def: maximum length}) and $p\leq T$.
\end{proof}

\subsection{Proof of Lemma~\ref{lem: uniform ix bound}}
\begin{proof}
  For each pair $(p,s)\in \sN^2$, define
\begin{equation*}
  \delta_{p,s} := \frac{\delta}{3p^2s^2}.
\end{equation*}
Since
\begin{equation*}
  \sum_{p=1}^\infty \sum_{s=1}^\infty \delta_{p,s} = \frac{\pi^4}{108}\delta < \delta,
\end{equation*}
applying Lemma~\ref{lem: thm 1 of neu} with $\delta_{p,s}$ and taking a union bound over all $(p,s)$ shows that
\begin{equation*}
  \Reg_{\mathrm{IX}}(s;p) \leq 4\sqrt{Ks\log K} + \qty(2\sqrt{\frac{Ks}{\log K}}+1) \log\qty(\frac{6p^2 s^2}{\delta})
\end{equation*}
for any realized pair $(p,s)$ with probability at least $1-\delta$.
For every recovery phase, we have $p\leq T$ and $s\leq |\gB_p^{\mathrm{ix}}\cap[T]| \leq L_T$, which concludes the proof.
\end{proof}

\section{Proof of Theorem~\ref{thm: sto cum}: stochastic regime}\label{app: sto}
In this section, we consider primarily the loss representation $\ell_{t,i}= 1- R_{t,i}$ to keep the notation consistent with our loss estimators.
Due to the similar phase-dependent policy structure, our analysis follows the outline of \citet{stahlbuhk2021learning}, although the technical arguments are different.

Recall that, in the stochastic regime, $A_t \in [0,1]$ and $R_t \in [0,1]^K$ are i.i.d.~across all rounds.
First, for each busy period $\gB_p$, we define the good-period indicator by
\begin{equation}\label{def: good period}
  \gJ_p := \I\qty[\hat{i}^*_p = i^*, \gB_p^{\mathrm{ix}}=\emptyset].
\end{equation}
In other words, $\gJ_p=1$ implies that the empirical best arm in exploration phases is correctly identified and the queue becomes empty without entering the recovery phase, and otherwise, $\gJ_p=0$.
Therefore, our policy always selects $i^*$ throughout the busy period $\gB_p$ if $\gJ_p=1$.
Then, we have the following lemmas, whose proofs are given in Appendix~\ref{app: sto lem}.
\begin{lemma}\label{lem: regret bad period}
  For $K\geq 2$ and $\epsilon = \mu_{i^*} - \lambda \in (0,1]$, Algorithm~\ref{alg: policy} satisfies that
  \begin{equation*}
    \E\qty[\sum_{t \in \gB_p}(Q_t^\pi - Q_t^{i^*})^+] \leq C_{p} \Pr(\gJ_p=0),
  \end{equation*}
  where
  \begin{equation*}
    C_p \leq c\qty(p\sqrt{p\log p} + \frac{p \log p}{\epsilon^2} + \frac{K^2 \log^2 K}{\epsilon^4})
  \end{equation*}
  for a universal constant $c>0$.
\end{lemma}
\begin{lemma}\label{lem: prob of bad period}
  For $p \geq \lceil 1+ \frac{2}{\epsilon} \rceil$, it holds that 
  \begin{equation*}
    \Pr(\gJ_p=0) \leq K\exp(-\frac{\Delta_{\min}^2 p\rho }{8K}) + (p+1)^{-7} + \exp(-\frac{\epsilon^2 p}{8}).
  \end{equation*}
\end{lemma}
\begin{proof}[Proof of Theorem~\ref{thm: sto cum}]
  Recall that busy and empty periods are defined by the policy’s queue, so $Q_t^{i^*} >0$ is possible even during an empty period.
  On the good busy period $\gB_p$ where $\gJ_p=1$, Algorithm~\ref{alg: policy} selects $i^*$ throughout the exploitation phase, equivalent to whole busy period, including its first round $\beta_p$.
  Since $Q_{\beta_p}^\pi = 0 \leq Q_{\beta_p}^{i^*}$, monotonicity of the queue recursion under the same arrivals and services imply that
  \begin{align}\label{eq: reason why arrival observation}
    Q_t^\pi \leq Q_t^{i^*}, \forall t \in \gB_p, \forall p: \gJ_p=1 \implies \sum_{t \in \gB_p} (Q_t^\pi - Q_t^{i^*})^+ = 0.
  \end{align}
  Establishing this result for our policy requires the assumption that arrivals are observed before decision, as explained in Appendix~\ref{app: why arrival}.
  Therefore, if $\gJ_p=1$, $\gB_p$ contribute no positive cumulative queue length regret so that considering the busy period with $\gJ_p=0$ is sufficient.

  Since $Q_t^\pi =0$ in the empty period, we have
    \begin{align*}
        \sum_{t=1}^\infty \E\qty[\qty(Q_t^\pi - Q_t^{i^*})] \leq  \sum_{p=1}^\infty \E\qty[\sum_{t\in \gB_p}\qty(Q_t^\pi - Q_t^{i^*})].
    \end{align*}
  By Lemma~\ref{lem: regret bad period}, there exists a constant $c >0$ satisfying for all $p\geq 1$
  \begin{equation*}
    \E\qty[\sum_{t \in \gB_p}(Q_t^\pi - Q_t^{i^*})] \leq c \qty(p\sqrt{p\log p} + \frac{p \log p}{\epsilon^2} + \frac{K^2 \log^2 K}{\epsilon^4})\Pr(\gJ_p=0).
  \end{equation*}
  Let $p_\Delta = \lceil 1+ \frac{2}{\epsilon} \rceil$.
  Then, by Lemma~\ref{lem: prob of bad period} and $\log p \leq p$, it suffices to show that there exists a finite constant $C_{\mathrm{sto}}$ satisfying
  \begin{multline*}
      \sum_{p= p_\Delta}^\infty c\qty((1+\epsilon^{-2})p^2 + \frac{K^2 \log^2 K}{\epsilon^4}) \qty(K\exp(-\frac{\Delta_{\min}^2p\rho}{8K})+(p+1)^{-7} + \exp(-\frac{\epsilon^2 p}{8})) \\
      + \sum_{p=1}^{p_\Delta} c\qty((1+\epsilon^{-2})p^2 + \frac{K^2 \log^2 K}{\epsilon^4}) \leq C_{\mathrm{sto}}.
  \end{multline*}
  For simplicity, let
  \begin{equation*}
    a=\frac{\Delta_{\min}^2 \rho}{8K}, \, b= \frac{\epsilon^2}{8}, \, C' = \frac{K^{2}\log^2 K}{\epsilon^4}.
  \end{equation*}
  \paragraph{First term.} Consider the following term,
  \begin{align*}
        \sum_{p= p_\Delta}^\infty cK ((1+\epsilon^{-2})p^2+C')\exp(-a\sqrt{p}).
    \end{align*}
    Then, by simple arithmetic calculation, we have
    \begin{align*}
        \sum_{p= p_\Delta}^\infty cK((1+\epsilon^{-2})p^2+C')e^{-ap} &\leq ce^a\int_{p_\Delta}^\infty( (1+\epsilon^{-2})x^2+C')e^{-ax} \dd x\\
        &\leq cKe^a \qty(\frac{2(1+\epsilon^{-2})}{a^3}+ \frac{C'}{a}) \\
        &\leq \gO\qty(K\exp(\frac{\Delta_{\min}^2 \rho}{K})\qty(\frac{K^3}{\epsilon^2 \Delta_{\min}^6 \rho^3} + \frac{K^3 \log^2 K}{\epsilon^4 \Delta_{\min}^2 \rho}))\\
        &= \gO\qty(\frac{K^4}{\epsilon^2 \Delta_{\min}^6 \rho^3} + \frac{K^4 \log^2 K}{\epsilon^4 \Delta_{\min}^2 \rho}).
    \end{align*}
    \paragraph{Second term.}
    The second part is indeed negligible since $p^{-7}$ is dominating, i.e.,
    \begin{equation*}
        \sum_{p=p_\Delta}^\infty c((1+\epsilon^{-2})p^2+C')(p+1)^{-7} \leq \gO(1+\epsilon^{-2} + C').
    \end{equation*}
    \paragraph{Third term.}
    Similarly to the first term, we obtain
    \begin{align*}
        \sum_{p= p_\Delta}^\infty c((1+\epsilon^{-2})p^2+C')e^{-bp} &\leq ce^b\int_{p_\Delta}^\infty( (1+\epsilon^{-2})x^2+C')e^{-bx} \dd x\\
        &\leq ce^b \qty(\frac{2(1+\epsilon^{-2})}{b^3}+ \frac{C'}{b}) \\
        &\leq \gO\qty(\epsilon^{-8}+C'\epsilon^{-2}) \tag{$\because \epsilon \in (0,1)$}
    \end{align*}
    \paragraph{Fourth term.}
    Finally, we have
    \begin{align*}
        \sum_{p=1}^{p_\Delta} c\qty((1+\epsilon^{-2})p^2 + \frac{K^2 \log^2 K}{\epsilon^4}) &\leq \gO\qty(\epsilon^{-2}p_{\Delta}^3 + \frac{K^2 \log K p_\Delta}{\epsilon^{4}}) \\
        &\leq \gO\qty(\epsilon^{-2}\Delta_{\min}^{-6} + \frac{K^2 \log K}{\Delta_{\min}^2 \epsilon^{4}}).
    \end{align*}
    \paragraph{Conclusion.}
    In sum, by definition of $p_{\Delta}$, we obtain
    \begin{align*}
        \sum_{p=1}^\infty \E\qty[\sum_{t\in \gB_p}\qty(Q_t^\pi - Q_t^{i^*})_+] \leq C_{\mathrm{sto}} = \mathrm{poly}\qty(K, \Delta_{\min}^{-1}, \epsilon^{-1}, \log K),
    \end{align*}
    which is finite since $\epsilon >0$, $\Delta_{\min} >0$, and $K < \infty$.

    Therefore, for any $T\geq1$,
    \begin{align*}
        \Reg_\pi^{\mathrm{cum}}(T) &=  \sum_{t=1}^T \E\qty[Q_t^\pi-Q_t^{i^*}] \leq \sum_{t=1}^\infty \E\qty[(Q_t^\pi-Q_t^{i^*})^+] \leq C_{\mathrm{sto}},
    \end{align*}
    which concludes the proof.
\end{proof}

\subsection{Observation of arrivals}\label{app: why arrival}
Observing $A_t$ before selecting $i_t$ allows the policy to distinguish an opportunity for exploration from the first round of a new busy period. 
This distinction prevents the first decision of a good busy period from inducing excess backlog relative to the comparator and supports the queue comparison used in our stochastic analysis.

Let $\beta_p$ be the first round of the busy period $\gB_p$.
By definition,
\begin{equation*}
  Q_{\beta_p}^\pi =0, \qq{and} A_{\beta_p} > 0.
\end{equation*}
Therefore, on the event $\gJ_p=1$, Algorithm~\ref{alg: policy} selects the optimal arm $i^*$ throughout $\gB_p$, including the round $\beta_p$.
Since busy and empty periods are defined by the policy’s queue, $Q_t^{i^*}$ need not be empty at $\beta_p$, which implies $Q_{\beta_p}^\pi \leq Q_{\beta_p}^{i^*}$.
Since both queues subsequently receive the same arrivals and service, monotonicity of the queue recursion preserves this ordering.
Therefore, 
\begin{equation*}
  \gJ_p = 1 \implies Q_t^\pi \leq Q_t^{i^*}, \forall t \in \gB_p,
\end{equation*}
which implies
\begin{equation*}
  \sum_{t \in \gB_p} (Q_t^\pi - Q_t^{i^*}) \leq 0.
\end{equation*}
Thus, these good busy periods contribute no positive cumulative queue length regret.

This argument relies on selecting $i^*$ from the first round of the busy period, where $Q_t^\pi=0$ and $A_t>0$.
To see what can fail without arrival observation, consider the modification $\bar{\pi}$ that explores uniformly whenever $Q_t^{\bar\pi}=0$, before observing $A_t$.
If $A_t=0$, exploration does not change the queue status regardless of the selected arm. 
If $A_t>0$, however, the decision before the newly arrived data, giving
\begin{equation*}
  Q_{t+1}^{\bar{\pi}} = (A_t - R_{t,i_t})^+ \geq 0.
\end{equation*}
In particular, if $Q_t^{i^*}=0$ as well, then
\begin{align*}
  Q_{t+1}^{\bar{\pi}}- Q_{t+1}^{i^*}  = (A_t - R_{t,i_t})^+ - (A_t - R_{t,i^*})^+,
\end{align*}
which can be positive.
The queue comparison can therefore incur linear regret even when the empirical best arm is already correct and the policy selects it in every subsequent round. 
Although this discrepancy may disappear after the queue becomes empty, similar costs can incur at the start of later busy periods.

Indeed, this modification can incur linear cumulative regret on an instance satisfying our stochastic assumptions. 
Consider an instance
\begin{equation*}
  K=2,\quad A_t \sim \mathrm{Bernoulli}(1/2), \quad R_{t,1} = 1, \quad R_{t,2} = 0, \forall t.
\end{equation*}
Here, $\rho=1/2$, arrivals and services are independent, and $i^*=1$ is uniquely optimal and stabilizing. 
The comparator queue satisfies $Q_t^{i^*}=0$ for every $t$.

When $Q_t^{\bar{\pi}}=0$, uniform exploration incurs one unit of backlog if $A_t=1$ and arm $2$ is selected, an event with conditional probability $1/4$.
When $Q_t^{\bar{\pi}}>0$, the queue is integer-valued and therefore at least one.
On the event $A_t=1$, service of at most one unit cannot empty this queue, regardless of the selected arm. 
Thus,
\begin{equation*}
  \E\qty[Q_{t+1}^{\bar{\pi}} \middle | Q_{t}^{\bar{\pi}}=q] \geq \begin{cases}
    1/4, & q=0, \\
    1/2, & q\geq 1.
  \end{cases}
\end{equation*}
This implies that for every $T\geq 1$, we have
\begin{equation*}
  \E\qty[\sum_{t=1}^T \qty(Q_t^{\bar{\pi}} - Q_t^{i^*})] \geq \frac{T-1}{4}.
\end{equation*}
Therefore, uniform exploration based only on the pre-arrival condition $Q_t^{\bar\pi}=0$ cannot achieve a sublinear stochastic cumulative-regret guarantee in our queue recursion definition.

The corresponding issue does not arise under the queue dynamics of \citet{stahlbuhk2021learning}, where they use
\begin{equation*}
  Q_{t+1} = (Q_t- R_{t,i_t})^+ + A_t.
\end{equation*}
In their model, arrivals in round $t$ become available for service only in the following round.
Therefore,
\begin{equation*}
  Q_t^\pi =0 \implies Q_{t+1}^\pi = A_t \leq (Q_{t}^{i^*}-R_{t,i^*})^+ + A_t = Q_{t+1}^{i^*}
\end{equation*}
holds regardless of the decision in round $t$.
Under their queue dynamics, exploration during empty periods preserves this queue comparison.

In other words, our arrivals-first assumption is introduced since we adopt the standard Lindley recursion
\begin{equation*}
  Q_{t+1}^\pi = (Q_t^\pi + A_t - R_{t,i_t})^+, 
\end{equation*}
which is also used in prior work on stochastic and adversarial queueing bandits~\citep{krishnasamy2016regret,freund2024efficient, kim2024queueing, krishnakumar2025minimizing, bae2026queue}.
Under this formulation, newly arriving data can receive service in the same round. 

Our additional assumption that arrivals are observed before decision allows the policy to identify rounds in which exploration cannot create backlog.
Specifically, the policy explores when $Q_t^\pi+A_t=0$ and begins exploitation immediately when $Q_t^\pi=0$ and $A_t>0$.
Only the indicator $\I[Q_t^\pi+A_t=0]$ is needed before the decision and the exact arrival amount can be revealed afterward for the queue updates.

\textbf{Remark 2 (Adversarial guarantee without arrival observation).} 
Consider a variant of Algorithm~\ref{alg: policy} that explores whenever $Q_t^\pi=0$ and starts exploitation at the first round with $Q_t^\pi>0$, i.e., where its busy periods are maximal intervals on which $Q_t^\pi>0$.
Let $\tilde{L}_T$ denote the longest such period up to $T$.
The bound of Theorem~\ref{thm: adv} holds for this variant with $L_T$ replaced by $\tilde{L}_T$.
The variant does not need to observe $A_t$, since its auxiliary queue $O_t$ can be updated from consecutive queue lengths and the observed service of the selected channel. 
Thus, arrival observation is required for our stochastic cumulative-regret guarantee, but not for the adversarial guarantee.

\section{Proofs of Lemmas for Theorem~\ref{thm: sto cum}}\label{app: sto lem}

\subsection{Proof of Lemma~\ref{lem: regret bad period}}
To prove this lemma, we first need the following result, whose proof is given in Appendix~\ref{app: aux lem}.
\begin{lemma}\label{lem: recovery round bounds}
  Let $\tau_q := \inf\{s\geq 1 : Q_{\tau(p,s)}^\pi =0, Q_{\tau(p,1)}=q \}$ denote the total number of local recovery rounds with the first round queue length $q>0$.
  Then, for any $K\geq 2$ and $\epsilon = \mu_{i^*} - \lambda  \in (0,1]$, it holds that 
  \begin{align*}
    \Pr(\tau_q > s) \leq \exp(-\frac{\epsilon^2 s}{18}) + 2\exp(-\frac{\epsilon}{18}\sqrt{\frac{s\log K}{K}}).
  \end{align*}
  for $s \geq s_0(q) = \max(\frac{3q}{\epsilon}, \frac{576K\log K}{\epsilon^2})$.
  Therefore, $\tau_q$ has a finite $r$-th moment satisfying
    \begin{equation*}
        \E[\tau_q^r] \leq (s_0(q))^r + e^{1/18}r\Gamma(r) \qty(\frac{18}{\epsilon^2})^r + 4e^{1/18}r\Gamma(2r)\qty(\frac{364K}{\epsilon^2 \log K})^r,
    \end{equation*}
    which implies that every fixed moment is polynomially bounded in terms of $q,K$ and $\epsilon$.
\end{lemma}
\begin{proof}
  Since $\sum_{t\in \gB_p} (Q_t^{\pi} - Q_t^{i^*})_+=0$ if $\gB_p$ satisfies $\gJ_p=1$ as in (\ref{eq: reason why arrival observation}), it suffices to consider the case of $\gJ_p=0$.
  Hence, we consider the exploitation phase and recovery phase, respectively.

  \paragraph{Exploitation phase.}
    Note that $O_t=0$ and thus $E_t=Q_t^\pi$ in the stochastic setting with $A_t \in [0,1]$ by (\ref{def: Et}).
    Therefore, at the $n$-th round of exploitation phase in busy period $\gB_p$, threshold condition implies that
    \begin{equation*}
        Q_t^\pi \leq g_p(n) = 4\sqrt{n\log(p+1)}.
    \end{equation*}
    Since the exploitation phase occurs at most $p$ rounds in $\gB_p$ by time-out condition, it holds
    \begin{align*}
        \sum_{t \in \gB_{p}^{\mathrm{ex}} } (Q_t^{\pi} - Q_t^{i^*})_+ &\leq \sum_{n=1}^p 4\sqrt{n\log(p+1)} \\
        &\leq 4p^{3/2}\sqrt{\log(p+1)}.
    \end{align*}
    If it does not transition to the recovery phase, $\gB_{p}^{\mathrm{ex}}=\gB_p$ holds, which concludes the proof.

  \paragraph{Recovery phase.}
  Therefore, we consider the case where the recovery phase exists.
  Let $Q_{\tau(p,1)}^\pi=q$, which implies that the exploitation ends with queue length $q>0$ and thus satisfies
  \begin{equation}\label{eq: q bound}
    q \leq g_p(n) + 1\leq 4\sqrt{n\log(p+1)}+1.
  \end{equation}
  For simplicity, we index the recovery phase from 1 so that $Q_0^\pi=q$.
  Since $A_t \in [0,1]$, it holds that
  \begin{align*}
      \sum_{s=0}^{\tau_q -1} (Q_s^{\pi} - Q_s^{i^*})_+ &\leq \sum_{s=0}^{\tau_q -1} (q+s) \\
      &\leq q\tau_q + \frac{\tau_q^2}{2}.
  \end{align*}
  By applying Lemma~\ref{lem: recovery round bounds} for $r=1,2$, we obtain
  \begin{align*}
      \E\qty[\sum_{s=0}^{\tau_q -1} (Q_s^{\pi} - Q_s^{i^*})_+] &\leq \gO\qty(qs_0(q)+(s_0(q))^2+ \frac{1}{\epsilon^2} + \frac{K}{\epsilon^2 \log K} + \frac{1}{\epsilon^4} + \frac{K^2}{\epsilon^4 \log^2 K}) \\
      &\leq \gO\qty(\frac{q^2}{\epsilon} +\frac{q K\log K}{\epsilon^2}+\frac{q^2}{\epsilon^2} + \frac{K^2 \log^2 K}{\epsilon^4}  + \frac{K}{\epsilon^2 \log K} + \frac{1}{\epsilon^4} + \frac{K^2}{\epsilon^4 \log^2 K}) \\
      &\leq \gO\qty(\frac{q^2}{\epsilon^2} + \frac{qK\log K}{\epsilon^2} + \frac{K^2 \log^2 K}{\epsilon^4}) \\
      &\leq \gO\qty(\frac{q^2}{\epsilon^2} + \frac{K^2 \log^2 K}{\epsilon^4}) \tag{$\because ab \leq (a^2+ b^2)/2$} \\
      &\leq \gO\qty(\frac{p\log p}{\epsilon^2} + \frac{K^2\log^2 K}{\epsilon^4}), \tag{by (\ref{eq: q bound})}
  \end{align*}
  which concludes the proof.
\end{proof}

\subsection{Proof of Lemma~\ref{lem: prob of bad period}}
The following results are required to provide the upper bounds on the probability of $\gJ_p=0$, whose proofs are given in Appendix~\ref{app: aux lem}.
\begin{lemma}\label{lem: error best arm}
  For any $m_0 \geq 1$, it holds that
  \begin{equation*}
    \Pr(\exists m \geq m_0: \argmin_{i \in [K]} \hat{L}^{\emptyset}_{m+1,i} \ne i^*) \leq (K-1)\exp(-\frac{m_0 \Delta_{\min}^2}{4K}).
  \end{equation*}
\end{lemma}
Note that by our index notation, $\hat{L}_{1,i}^{\emptyset} = 0$, and thus the probability is $1-1/K$ for $m=0$ case.
\begin{proof}[Proof of Lemma~\ref{lem: prob of bad period}]
  By definition, $\gJ_p=0$ if its empirical best arm is incorrect or if it enters the recovery phase despite identifying the correct best arm.
  Let
  \begin{equation*}
    m_p := \sum_{j=1}^p  |\gE_j|
  \end{equation*}
  be the number of exploration rounds up to the beginning of busy period $\gB_p$.
  Since the $\gE_j$ can be empty due to positive arrivals right after the queue becomes empty, let $\alpha_j$ be the first round with $Q_t^\pi =0$ after the ($j-1$)-th the busy period, with $\alpha_1=1$. 
  \begin{equation*}
    Z_j := \I[A_{\alpha_j} = 0].
  \end{equation*}
  Therefore, $Z_j=1$ implies that Algorithm~\ref{alg: policy} is in exploration phase at the round $\alpha_j$, which implies
  \begin{equation*}
    m_p \geq \sum_{j=1}^p Z_j.
  \end{equation*} 
  Since $\alpha_j$ is determined by the history before its arrival, i.i.d.~arrival assumption implies that $Z_j$ are independent Bernoulli random variables with success probability $\rho$.
  Hence, its summation follows the Binomial distribution with $p$ trials and $\rho$ success probability.
  By applying multiplicative Chernoff bound, we obtain
  \begin{align}\label{eq: mp prob bound}
    \Pr(m_p < \frac{\rho p}{2}) \leq \exp(-\frac{\rho p}{8}).
  \end{align}
  \paragraph{Incorrect empirical best arm.}
  By the definition of $i^*_p$ in (\ref{def: empirical best}), Lemma~\ref{lem: error best arm} implies that for $m^* = \rho p /2$
  \begin{align*}
    \Pr(i^*_p \ne i^*) &\leq \Pr(m_p < m^*) +\Pr(\argmin_{i\in [K]} \hat{L}_{m+1,i}^{\emptyset} \ne i^*, m_p \geq m^*) \\
    &\leq \exp(-\frac{\rho p}{8}) + (K-1) \exp(-\frac{\Delta_{\min}^2 \rho p}{8K}) \\
    &\leq K \exp(-\frac{\Delta_{\min}^2 \rho p}{8K}).  \tag{$\because \Delta_{\min} \in (0,1], K \geq 2$}
  \end{align*}
  \paragraph{Recovery phase with correct best arm.}
  Let us consider the case where $i^*_p = i^*$ holds.
  Then, during the exploitation phase $\gB_p^{\mathrm{ex}}$, the policy always selects $i^*$.
  Define the residual queue length by $\bar{Q}_{s} := A_s - R_{s,i^*}$, which satisfies
  \begin{equation*}
    \bar{Q}_s \in [-1,1], \quad \E[\bar{Q}_s] = \lambda - \mu_{i^*} = -\epsilon.
  \end{equation*}
  Therefore, $\bar{Q}_s+ \epsilon$ is of expectation zero.
  \paragraph{Case 1: threshold violation.}
  Consider the case the exploitation phase end because the queue length exceeds the threshold.
  Notice that the threshold violation cannot occur at local exploitation round $n=1$ since $Q_{\beta_p}^\pi = 0$ and at $n=2$ since $Q_{\beta_p +1}^\pi \leq 1 < g_p(2)$ by definitions.

  If such threshold violation occurs at round $n\geq 3$, this means that the preceding $n-1$ times selecting $i^*$ did not empty the queue.
  Therefore,
  \begin{align*}
    Q_{\beta_p + n -1}^\pi = A_{\beta_p} - R_{\beta_p, i^*} + \sum_{j=1}^{n-2} \bar{Q}_j \leq 1 +  \sum_{j=1}^{n-2} \bar{Q}_j .
  \end{align*}
  By applying Hoeffding's inequality, we obtain
  \begin{align*}
    \Pr(Q_{\beta_p + n -1}^\pi > g_p(n)) &\leq \Pr\qty(\sum_{j=1}^{n-2}(\bar{Q}_s + \epsilon) > g_p(n)-1 + \epsilon(n-2)) \\
    &\leq \exp(-\frac{(g_p(n)-1)^2}{n-2}) \leq \exp(-\frac{9n\log(p+1)}{n-2}) \leq (p+1)^{-8}.
  \end{align*}
  Since such threshold violation occur at most once for each busy period, a union bound gives
  \begin{equation*}
    \Pr(\mathrm{threshold~violation~recovery}, \hat{i}^*_p = i^*) \leq p(p+1)^{-8} \leq (p+1)^{-7}.
  \end{equation*}
  \paragraph{Case 2. time-out.}
  If the recovery phase starts by the time-out after $p$ optimal arm selections, the queue remains positive after these $p$ rounds.
  Therefore, 
  \begin{equation*}
    0 < Q_{\beta_p + p}^\pi  \leq 1 + \sum_{j=1}^{p-1} \bar{Q}_j \implies \sum_{j=1}^{p-1} (\bar{Q}_j + \epsilon) > \epsilon(p-1)-1.
  \end{equation*}
  Here, for $p\geq 1 + 2/\epsilon$, $\epsilon(p-1)-1 \geq \epsilon(p-1)/2$ holds.
  By applying Hoeffding's inequality, we obtain
  \begin{equation*}
    \Pr(\text{time-out~recovery}, \hat{i}^*_p = i^*) \leq \exp(-\frac{\epsilon^2 (p-1)}{4}) \leq \exp(-\frac{\epsilon^2 p}{8}).
  \end{equation*}
  Finally, combining all the results with a union bound concludes the proof.
\end{proof}

\section{Proof of Corollary~\ref{cor}}\label{app: cor}
\begin{proof}
  By definition, we have
  \begin{align*}
      Q_t^\pi -\min_{i\in [K]} Q_t^i \leq  Q_t^\pi -Q_t^{i^*} +  Q_t^{i^*} -\min_{i\in [K]} Q_t^i,
  \end{align*}
  which implies
  \begin{align*}
      \Reg_{\pi}^{\max}(T) &\leq \max_{t\in [T]} \qty( Q_t^\pi -Q_t^{i^*})^+ + \max_{t\in [T]} \qty( Q_t^{i^*} -\min_{i\in [K]} Q_t^i)^+.
  \end{align*}
  By Theorem~\ref{thm: sto cum}, the first term satisfies
    \begin{equation*}
        \E\qty[\max_{t\in [T]} \qty( Q_t^\pi -Q_t^{i^*})_+] \leq \sum_{t=1}^\infty \E\qty[\qty( Q_t^\pi -Q_t^{i^*})_+] \leq C_{\mathrm{sto}}.
    \end{equation*}
    For the second term, let us consider the busy period $\gB_p^*$ in terms of constant policy of selecting $i^*$, with its initial round $\beta_p^*$.
    Then, for $t\in \gB_p^*$, we have
    \begin{align*}
      Q_t^{i^*} = \sum_{s=\beta_p^*}^{t-1} (A_s - R_{s,i^*}).
    \end{align*}
    During this period, for any $j\ne i^*$, it holds that
    \begin{align*}
      Q_{t}^j \geq \sum_{s=\beta_p^*}^{t-1} (A_s -R_{s,j}). 
    \end{align*}
    Hence, we obtain
    \begin{align*}
      Q_t^{i^*} - Q_t^{j} \leq \sum_{s=\beta_p^*}^{t-1} (R_{s,j} - R_{s,i^*}) \leq \max_{t' \in [t-1]} \sum_{s=t'}^{t-1} (R_{s,j} - R_{s,i^*}).
    \end{align*}
    Let $g_{s}^j = R_{s,j} - R_{s,i^*}$, which satisfies $g_{s}^j \in [-1,1]$ and $\E[g_s^j] = -\Delta_j$.
    For any $n \in \sN$, Hoeffding's inequality gives that
    \begin{align*}
      \Pr(\sum_{s=1}^n g_s^j \geq x) = \Pr(\sum_{s=1}^n (g_s^j + \Delta_j)\geq x+n\Delta_j) &\leq \exp(-\frac{(x+n\Delta_j)^2}{2n}) \\
        &\leq e^{-2x\Delta_j}. \tag{$\because (a+b)^2\geq 4ab$}
    \end{align*}
    By taking a union bound, we have for any $x>0$ that
    \begin{align*}
      \Pr(\max_{j\ne i^*} \max_{1\leq t' \leq t \leq T} \sum_{s=t'}^t g_s^j \geq x)  \leq \min\qty(1, (K-1)T^2 e^{-2x\Delta_{\min}}). 
    \end{align*}
    This is because the possible choices of interval pairs $[t',t] \subseteq [T]$ is $T(T+1)/2 \leq T^2$.
    Let $x_0 = \frac{\log((K-1)T^2)}{2\Delta_{\min}}$ so that $(K-1)T^2e^{-2x\Delta_{\min}} =1$.
    Since $E[X] = \int_0^\infty Pr(X\geq x) \dd x$ for any non-negative random variable $X$, we obtain
    \begin{align*}
        \E\qty[\max_{t \in [T]}\qty(Q_{t}^{i^*}-\min_{i\ne i^*} Q_t^i)^+] &\leq \int_0^\infty \min\qty(1, (K-1)T^2 e^{-2x\Delta_{\min}}) \dd x \\
        &\leq \int_0^{x_0}1 \dd x + \int_{x_0}^\infty (K-1)T^2 e^{-2x\Delta_{\min}} \dd x \\
        &= x_0 + \frac{(K-1)T^2e^{-2x_0\Delta_{\min}}}{2\Delta_{\min}} \\
        &= x_0 + \frac{1}{2\Delta_{\min}}, \tag{by definition of $x_0$}
    \end{align*}
    which concludes the proof.
\end{proof}

\section{Proofs of auxiliary lemmas}\label{app: aux lem}

\subsection{Proof of Lemma~\ref{lem: Q,E,g}}
\begin{proof}
  Indeed, the lemma follows directly from the construction of the policy and the relations between $Q_t^\pi$, $O_t$, and $E_t$.
  Fix any $t\in\gB_p^{\mathrm{ex}}$, and let $n\geq1$ be its local exploitation round. 
  Since the policy remains in the exploitation phase at round $t$, the threshold condition in (\ref{def: threshold}) ensures that
  \begin{equation*}
    Q_t^\pi - O_t = E_t \leq g_p(n) = 4\sqrt{n \log(p+1)}.
  \end{equation*}
  Combining this inequality with (\ref{eq: E and Q}) concludes the first part.
  The second part follows from $E_t\leq E_{t-1}+1$.
\end{proof}

\subsection{Proof of Lemma~\ref{lem: Q,E,Reg}}
\begin{proof}
  Fix any $t\in \gB_p^{\mathrm{ix}}$, and let $s\geq1$ be its local recovery round, so that $t=\tau(p,s)$. 
  Since the policy queue remains non-empty throughout the busy period, its recursion is linear between the first recovery round and round $s$. 
  Therefore,
  \begin{equation*}
    Q_t^{\pi} = Q_{\tau(p,1)}^{\pi} + \sum_{s'=1}^{s-1} (A_{\tau(p,s')} - R_{\tau(p,s'),i_{\tau(p,s')}}),
  \end{equation*}
  where $i_{s'}$ are selected by the corresponding Exp3-IX instance.
  In contrast, the queue length under the constant policy can be empty during this interval. 
  Nevertheless, since $(x)^+ \geq x$, we have for any $i\in[K]$
  \begin{equation*}
    Q_t^i \geq Q_{\tau(p,1)}^i + \sum_{s'=1}^{s-1} (A_{\tau(p,s')} - R_{\tau(p,s'),i}).
  \end{equation*}
  Therefore, for any $i\in [K]$ and $t\in \gB_p^{\mathrm{ix}}$ satisfying $t=\tau(p,s)$, we obtain
  \begin{align*}
    Q_t^\pi - Q_t^i &\leq Q_{\tau(p,1)}^{\pi} - Q_{\tau(p,1)}^{i} + \sum_{s'=1}^{s-1} (R_{\tau(p,s'),i} - R_{\tau(p,s'),i_{\tau(p,s')}}) \\
    &\leq  Q_{\tau(p,1)}^{\pi} - O_{\tau(p,1)} + \sum_{s'=1}^{s-1} (R_{\tau(p,s'),i} - R_{\tau(p,s'),i_{\tau(p,s')}}) \tag{by (\ref{eq: O and Q})} \\
    &\leq E_{\tau(p,1)} + \sum_{s'=1}^{s-1} (R_{\tau(p,s'),i} - R_{\tau(p,s'),i_{\tau(p,s')}}) \\
    &\leq E_{\tau(p,1)} + \sum_{s'=1}^{s} (R_{\tau(p,s'),i} - R_{\tau(p,s'),i_{\tau(p,s')}}) +1 \tag{$\because R_t \in [0,1]^K$}
  \end{align*}
  which concludes the proof.
  Note that the additive constant $1$ is not essential and is included only to simplify the presentation by shifting the index.
\end{proof}

\subsection{Proof of Lemma~\ref{lem: recovery round bounds}}
\begin{proof}
    Let $\gH_0$ denote the complete history until the beginning of the recovery phase, i.e., $\gF_s$ for $s=\tau(p,1)-1$.
    Under i.i.d.~stochastic model, the future observations during the recovery phase are independent of $\gH_0$ and have the same distribution as the original process.

    For simplicity, we use the round index from $0$, so that $Q_0^\pi=q$ holds at the beginning of the recovery phase.
    On the event $\{\tau_q > n\}$, the queue remains strictly positive throughout the first $n$ recovery rounds, i.e., $Q_s^\pi >0$ for all $s \in [n]$, which implies
    \begin{equation*}
        Q_n^\pi = q + \sum_{s=1}^n A_s - R_{s,i_s} > 0.
    \end{equation*}
    Note that $i_s$ is selected by Exp3-IX policy and
    \begin{equation}\label{eq: tau q control 1}
        0 < q + \sum_{s=1}^n A_s - R_{s,i^*} + \sum_{s=1}^n R_{s,i^*} -  R_{s,i_s}.
    \end{equation}
    Define
    \begin{equation*}
        F_n := \sum_{s=1}^n (A_s - R_{s,i^*} + \epsilon) \qq{and} G_n := \sum_{s=1}^n (R_{s,i^*} - R_{s,i_s}),
    \end{equation*}
    so that (\ref{eq: tau q control 1}) implies $q-\epsilon n + F_n + G_n >0$.

    Suppose $n\geq \frac{3q}{\epsilon}$, which implies that $\epsilon n - q \geq \frac{2\epsilon n}{3}$.
    Then, (\ref{eq: tau q control 1}) implies that $F_n + G_n > \frac{2\epsilon n}{3}$ holds if $n \geq 3q/\epsilon$.
    In other words,
    \begin{equation*}
        \qty{\tau_q > n} \subseteq \qty{F_n \geq \frac{\epsilon n}{3}} \cup \qty{G_n \geq \frac{\epsilon n}{3}}.
    \end{equation*}
    \paragraph{Case 1.} We first control the first event related to $F_n$.
    Let $f_s := A_s -R_{s,i^*}+\epsilon$, so that $F_n = \sum_{s=1}^n f_s$.
    By the definition of $\epsilon$, we have $\E[f_s]=0$ and $f_s \in [-1+\epsilon,1+\epsilon]$.
    Therefore, Hoeffding's inequality gives
    \begin{align*}
        \Pr(F_n > \frac{\epsilon n}{3}) \leq \exp(-\frac{\epsilon^2 n}{18}).
    \end{align*}
    \paragraph{Case 2.} Here, let us consider the case for $G_n$.
    By definition of $\Reg_{\mathrm{IX}}(n;p)$ in (\ref{def: regret ix}), it holds that
    \begin{equation*}
        G_n \leq \max_{i \in [K]} \sum_{r=1}^n R_{r,i} - R_{r,i_r}  = \Reg_{\mathrm{IX}}(n;p)
    \end{equation*}
    regardless of $p\geq 1$.
    Therefore, by choosing $\delta := 2\exp(-\frac{\epsilon n}{6(2d_n+1)})$ in Lemma~\ref{lem: thm 1 of neu} for $d_n=\sqrt{Kn/\log K}$, we have
    \begin{align*}
        \Pr(G_n > 4d_n \log K+ \qty(2d_n+1)\log \frac{2}{\delta}) \leq \delta.
    \end{align*}
    Here, by the choice of $d_n$ and $\delta$, for $K\geq 2$ and $n\geq 1$, we have $2d_n +1 \leq 3d_n$ and thus 
    \begin{align*}
        \frac{\epsilon n}{6(2d_n+1)} \geq \frac{\epsilon n}{18 d_n}.
    \end{align*}
    Hence, if $n \geq \frac{576K\log K}{\epsilon^2}$, 
    \begin{equation*}
        \frac{\epsilon n}{18 d_n} = \frac{\epsilon\sqrt{n\log K}}{18\sqrt{K}} \geq \frac{4}{3}\log K \implies \delta \leq 2K^{-4/3} < 1,
    \end{equation*}
    which implies our choice of $\delta$ is valid.
    Moreover, for $n \geq \frac{576K\log K}{\epsilon^2}$, we have
    \begin{align*}
        4d_n \log K = 4\sqrt{Kn\log K} \leq \frac{\epsilon n}{6}.
    \end{align*}
    Since $\log (2/\delta)= \frac{\epsilon n}{6(2d_n +1)}$ by definition, we obtain
    \begin{align*}
        4d_n \log K+ \qty(2d_n+1)\log \frac{2}{\delta} \leq \frac{\epsilon n}{3}.
    \end{align*}
    Therefore,
    \begin{align*}
        \Pr(G_n > \frac{\epsilon n}{3}) \leq \Pr(\Reg_{\mathrm{IX}}(n) > \frac{\epsilon n}{3}) \leq 2\exp(-\frac{\epsilon n}{6(2d_n+1)}) \leq 2\exp(-\frac{\epsilon}{18}\sqrt{\frac{n \log K}{K}}),
    \end{align*}
    which concludes the proof for the first part.

    Set $a=\frac{\epsilon^2}{18}$ and $b=\frac{\epsilon}{18}\sqrt{\frac{\log K}{K}}$.
    Then, for any $n\geq n_0(q)$, the first part shows that $\Pr(\tau_q > n) \leq e^{-an}+2e^{-b\sqrt{n}}$.
    For any integer $r\geq 1$, it holds that
    \begin{align*}
        \E[\tau_q^r] &= \sum_{n=0}^{\infty} ((n+1)^r - n^r) \Pr(\tau_q > n) \\
        &=\sum_{n=0}^{n_0(q)-1} ((n+1)^r - n^r) \Pr(\tau_q > n) + \sum_{n=n_0(q)}^{\infty} ((n+1)^r - n^r) \Pr(\tau_q > n) \\
        &\leq n_0^r(q)+ \sum_{n=n_0(q)}^{\infty} ((n+1)^r - n^r) \Pr(\tau_q > n) \\
        &\leq n_0^r(q)+ \sum_{n=n_0(q)}^{\infty} r(n+1)^{r-1}(e^{-an}+2e^{-b\sqrt{n}}) \tag{by mean-value theorem}
    \end{align*}
    Since $n_0(q)\geq 1$,
    \begin{align*}
        \sum_{n=n_0(q)}^{\infty} r(n+1)^{r-1}(e^{-an}+2e^{-b\sqrt{n}})  &\leq \sum_{n=1}^{\infty} r(n+1)^{r-1}(e^{-an}+2e^{-b\sqrt{n}}) \\
        &=\sum_{n=2}^{\infty} r n^{r-1}(e^{-a(n-1)}+2e^{-b\sqrt{n-1}}) \\
        &\leq \sum_{n=2}^{\infty} r n^{r-1}(e^a e^{-an}+2e^b e^{-b\sqrt{n}}) 
    \end{align*}
    By definition of Gamma function $\Gamma$, we have
    \begin{align*}
        \sum_{n=2}^\infty n^{r-1}e^{-an} &\leq \int_1^\infty z^{r-1}e^{-az}\dd z \\
        &\leq \frac{\Gamma(r)}{a^r},
    \end{align*}
    and
    \begin{align*}
        \sum_{n=2}^\infty n^{r-1}e^{-b\sqrt{n}} &\leq \int_1^\infty z^{r-1}e^{-b\sqrt{z}}\dd z \\
        &\leq \frac{2\Gamma(2r)}{b^{2r}},
    \end{align*}
  which concludes the proof.
\end{proof}

\subsection{Proof of Lemma~\ref{lem: error best arm}}
We first introduce a variant of Freedman's inequality.
\begin{lemma}[Corollary 1(b) of \citet{howard2020time}]\label{lem: freedman}
  Suppose $(Z_t)_{t\geq 0}$ is a real-valued martingale adapted to filtration $(\gF_t)_t$ with $Z_0=0$ satisfying $ Z_t - Z_{t-1} \leq c$ almost surely for all $t\geq 1$.
  Define the predictable quadratic variation
  \begin{equation*}
      V_t := \sum_{s=1}^t \E\qty[(Z_s - Z_{s-1})^2|\gF_{s-1}].
  \end{equation*}
  Then, for every $x, k \geq 0$,
  \begin{equation}\label{eq: freedman howard}
      \Pr\qty(\exists t \geq 1 : Z_t \geq x+ \mathfrak{s}_p\qty(\frac{x}{k})(V_t - k)) \leq \exp(-\frac{x^2}{2(k+cx/3)}),
  \end{equation}
  where $\mathfrak{s}_P(u) = \frac{cu-\log(1+cu)}{c\log(1+cu)}$.
\end{lemma}
\begin{proof}
Let $Y_{m,i} = \hat{\ell}_{m,i}^{\emptyset} -  \hat{\ell}_{m,i^*}^{\emptyset}$ and define a martingale difference sequence
\begin{equation}\label{def: mds X}
  X_{m,i} := \hat{\ell}_{m,i}^{\emptyset} -  \hat{\ell}_{m,i^*}^{\emptyset} - \E\qty[ \hat{\ell}_{m,i}^{\emptyset} -  \hat{\ell}_{m,i^*}^{\emptyset} \middle | \gF_{m-1}] = Y_{m,i} - \E\qty[Y_{m,i} \middle |\gF_{m-1}],
\end{equation}
for the global history $\gF_{m-1}$ until $m$-th exploration rounds.
Since $ \hat{\ell}_{m,i}^{\emptyset}$ is IW loss estimator with the uniform exploration, we have
\begin{equation*}
  \E[Y_{m,i}| \gF_{m-1}] = \Delta_i, \quad |Y_{m,i}| \leq K, \qq{and} \E[Y_{m,i}^2 | \gF_{m-1}] = K\E[\ell_{m,i}^2 +\ell_{m,i^*}^2] \leq K \E[\ell_{m,i} +\ell_{m,i^*}] \leq K(2-\Delta_i),
\end{equation*}
which implies
\begin{equation}\label{eq: mds bound}
  |X_{m,i}| \leq K, \qq{and} \E\qty[X_{m,i}^2 \middle| \gF_{m-1}] \leq  K(2-\Delta_i), \, \forall m\geq 1, i\ne i^*.
\end{equation}
The last inequality follows since the variance is at most the second moment, i.e., $\E[X_{m,i}^2| \cdot] \leq \E[Y_{m,i}^2| \cdot]$.

By definition, it holds that
\begin{equation*}
  \qty{ \argmin_{i\in [K]} \hat{L}_{m+1,i}^{\emptyset} \ne i^*} \subseteq \qty{\exists i \ne i^*: \hat{L}_{m+1,i}^{\emptyset} - \hat{L}_{m+1,i^*}^{\emptyset} \leq 0 }.
\end{equation*}
Here, 
\begin{align*}
  \hat{L}_{m+1,i}^{\emptyset} - \hat{L}_{m+1,i^*}^{\emptyset} = \sum_{b=1}^{m} Y_{b,i} &= \sum_{b=1}^{m} \E\qty[Y_{b,i}\middle| \gF_{b-1}] + \sum_{b=1}^{m} X_{b,i} \\
  &= m\Delta_i + \sum_{b=1}^{m} X_{b,i},
\end{align*}
which implies
\begin{equation*}
  \qty{\exists i \ne i^*: \hat{L}_{m+1,i}^{\emptyset} - \hat{L}_{m+1,i^*}^{\emptyset} \leq 0 } = \qty{\sum_{b=1}^{m} X_{b,i} \leq -m\Delta_i}.
\end{equation*}
Recall that $X_{m,i}$ is a martingale difference sequence satisfying (\ref{eq: mds bound}).
Then, by applying the variation of Freedman's inequality in Lemma~\ref{lem: freedman}\footnote{Notice that $\mathfrak{s}_P(u) >0$ for $u>0$.
This implies that if $x>0$
\begin{equation*}
    \Pr(\exists t \geq 1: Z_t \geq x, V_t \leq m) \leq  \Pr\qty(\exists t \geq 1 : Z_t \geq x+ \mathfrak{s}_p\qty(\frac{x}{m})(V_t - m)).
\end{equation*}}, we obtain for $m_0\geq 1$ that
\begin{align*}
  \Pr(\exists m \geq m_0: \sum_{b=1}^{m} X_{b,i} \leq -m\Delta_i) \leq \exp(-\frac{m^2_0\Delta_i^2}{2(Km_0(2-\Delta_i)+Km_0\Delta_i/3)}) \leq \exp(-\frac{m_0\Delta_i^2}{4K}).
\end{align*}
Therefore, taking a union bound over the suboptimal arms conclude the proof. 
\end{proof}

\section{Details on experiments}\label{app: exp}
As briefly discussed in Sections~\ref{sec: prelim} and~\ref{sec: policy}, the definitions of empty and busy periods and the queue dynamics differ across the policies considered in our experiments.
Let $Q_t^j$ be the queue length under the arrival-first convention in (\ref{def: queue lindley}) and $\bar{Q}_t^j$ be that under the service-first convention in (\ref{def: queue stan}) for the same selection sequence $j$.

We first show by induction that
\begin{equation}\label{eq: queue dynamics difference}
  0\leq \bar{Q}_t^{j}-Q_t^{j}\leq 1
\end{equation}
for every $t$. 
The claim holds for any empty period and at $t = \beta_p$ since both queues are initialized at $t=1$.
Suppose that (\ref{eq: queue dynamics difference}) holds at round $t\in \gB_p$ for some $p\in \sN$ and let $R_t$ denote the service obtained by the common arm.
Then, we have
\begin{align*}
  Q_{t+1}^j = (Q_t^j + A_t - R_t)^+ \leq (Q_t^j-R_t)^+ + A_t \leq (\bar{Q}_{t}^j - R_t )^+ + A_t = \bar{Q}_{t+1}^j.
\end{align*}
Using the induction hypothesis and $R_t\leq1$, we also obtain
\begin{align*}
  \bar{Q}_{t+1}^j \leq (Q_t^j + 1 - R_t)^+ + A_t \leq (Q_t^j + A_t - R_t)^+ + 1 = Q_{t+1}^j + 1,
\end{align*}
which proves (\ref{eq: queue dynamics difference}).
Therefore, the worst-case queue length regrets obtained by the same selection sequence under the two recursions differ by at most $1$.

The above comparison holds when the action sequence is fixed across the two queue dynamics. 
Since queue-dependent policies may select different actions under different recursions, it does not guarantee that their worst-case regrets differ by at most $1$ when they are run separately. 
Nevertheless, the direct discrepancy between the two queue recursions is bounded at each round, and worst-case regret takes a maximum rather than accumulating differences over time. 
We therefore expect the choice of queue dynamics to have a limited effect on worst-case regret. 
In contrast, cumulative regret aggregates roundwise differences and can be substantially more sensitive to this modeling choice.

To follow the original formulations, Section~\ref{sec: exp} reports results obtained by implementing each policy with the queue dynamics and period definitions used in the corresponding paper.
For our policy, empty and busy periods are defined by $Q_t^\pi+A_t=0$ and $Q_t^\pi+A_t>0$, respectively. 
For the baseline policies, we retain the convention that empty and busy rounds satisfy $Q_t^\pi=0$ and $Q_t^\pi>0$, respectively. 
In particular, our policy, Q-ThS, and KS-OGD use the arrivals-first dynamics in~(\ref{def: queue lindley}), whereas SSM-3P uses the service-first recursion in~(\ref{def: queue stan}).
This choice reproduces each policy under its original model, although cumulative-regret comparisons involving SSM-3P may partly reflect the difference in queue dynamics.

\begin{figure*}[ht]
    \centering
    \begin{subfigure}[t]{0.43\textwidth}
        \centering
        \includegraphics[width=\linewidth]
        {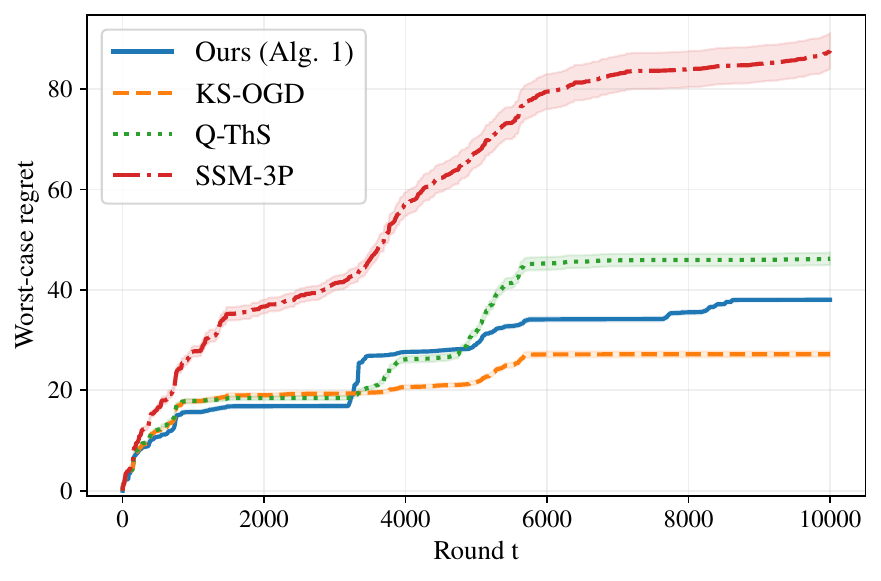}
        \caption{Worst-case regret in adversarial regime.}
        \label{fig:af ad worst}
    \end{subfigure}
    \hfil
        \begin{subfigure}[t]{0.43\textwidth}
        \centering
        \includegraphics[width=\linewidth]
        {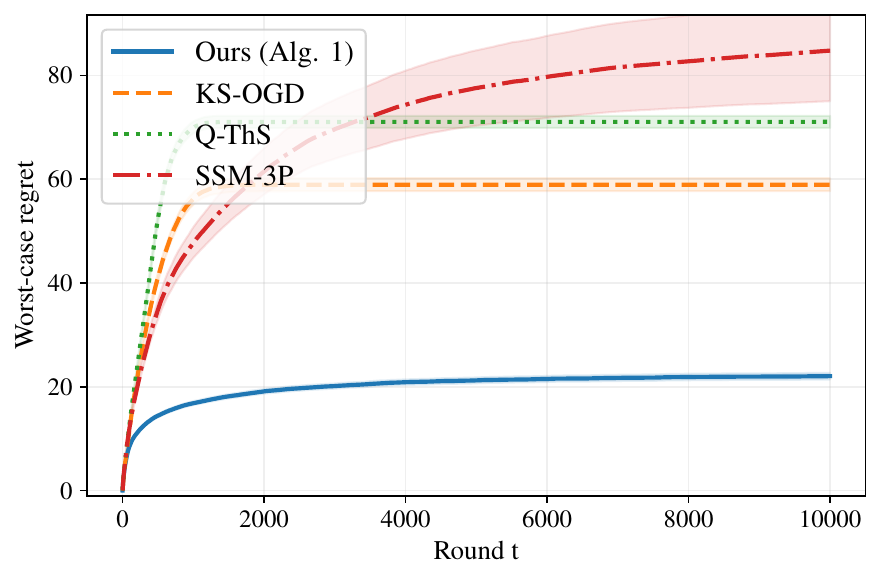}
        \caption{Worst-case regret in stochastic regime.}
        \label{fig:af st worst}
    \end{subfigure}
    \\
        \begin{subfigure}[t]{0.43\textwidth}
        \centering
        \includegraphics[width=\linewidth]
        {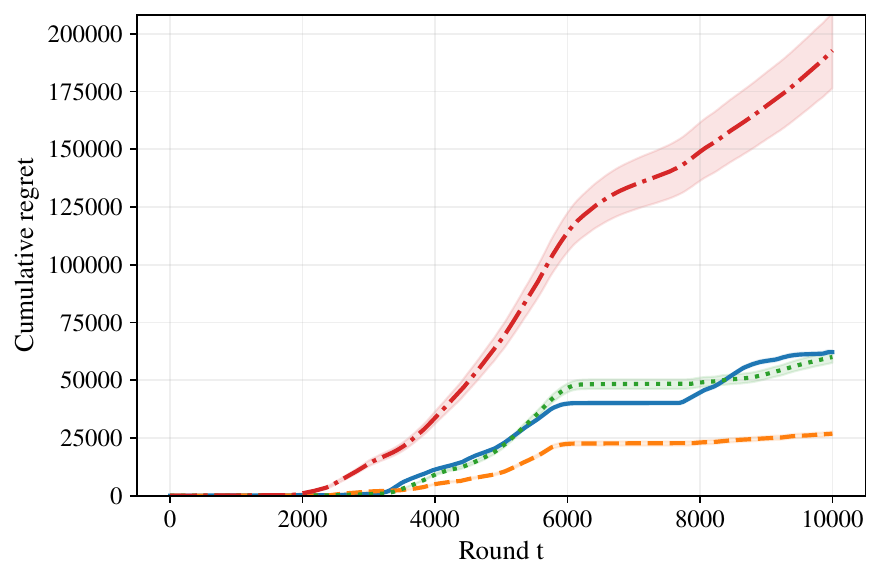}
        \caption{Cumulative regret in adversarial regime.}
        \label{fig:af ad cum}
    \end{subfigure}
    \hfil
    \begin{subfigure}[t]{0.43\textwidth}
        \centering
        \includegraphics[width=\linewidth]
        {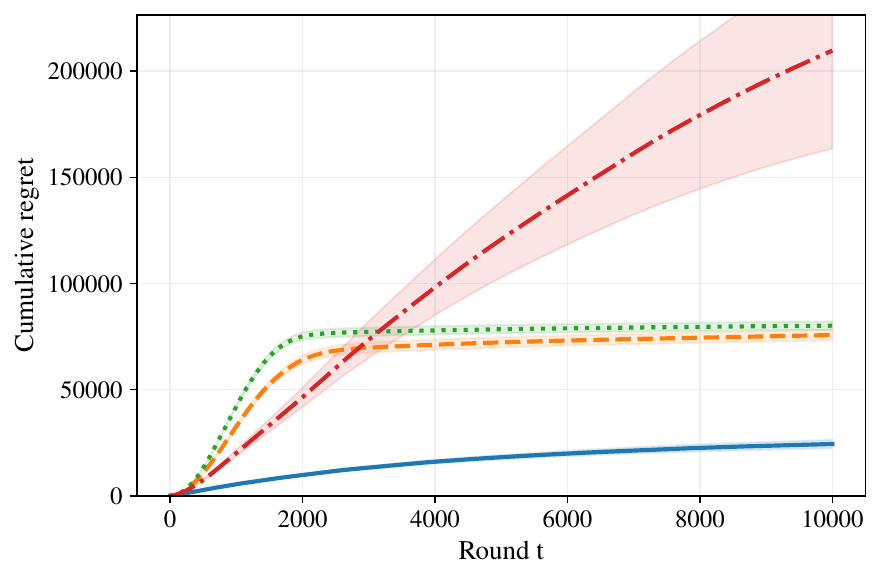}
        \caption{Cumulative regret in stochastic regime.}
        \label{fig:af st cum}
    \end{subfigure}
    \\
        \begin{subfigure}[t]{0.43\textwidth}
        \centering
        \includegraphics[width=\linewidth]
        {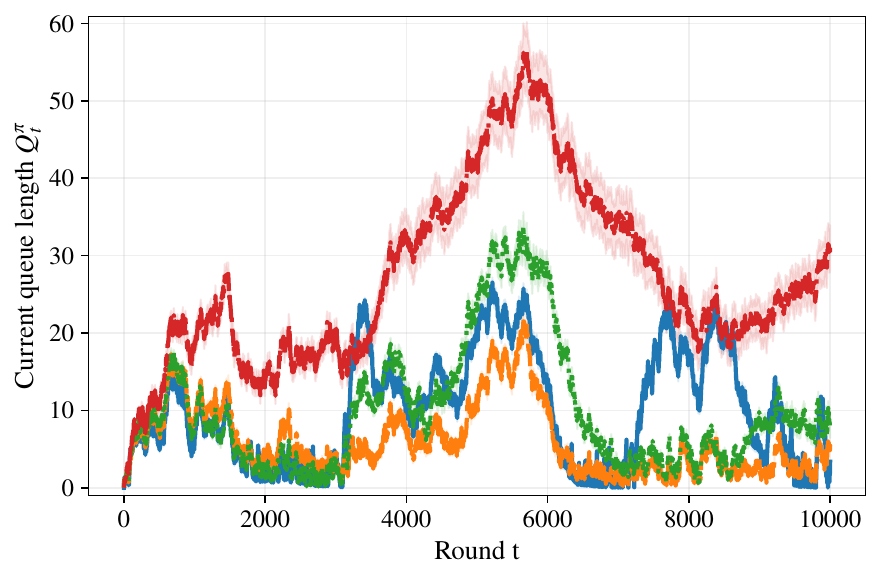}
        \caption{Queue lengths $Q_t^\pi$ in adversarial regime.}
        \label{fig:af ad queue}
    \end{subfigure}
    \hfil
    \begin{subfigure}[t]{0.43\textwidth}
        \centering
        \includegraphics[width=\linewidth]
        {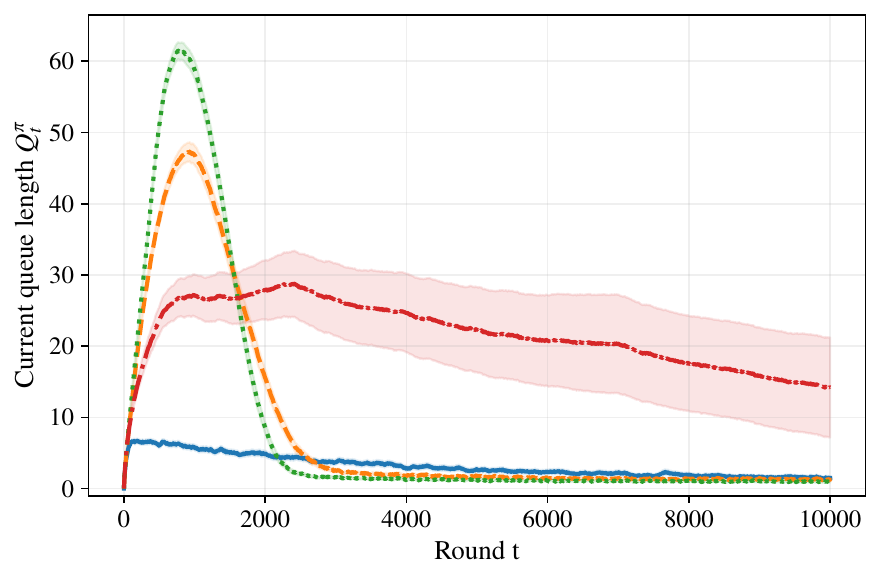}
        \caption{Queue lengths $Q_t^\pi$ in stochastic regime.}
        \label{fig:af st queue}
    \end{subfigure}
    \caption{Results under arrival-first dynamics in (\ref{def: queue lindley})} \vspace{-0.5em}
    \label{fig: af rslt}
\end{figure*}

\begin{figure*}[ht]
    \centering
    \begin{subfigure}[t]{0.43\textwidth}
        \centering
        \includegraphics[width=\linewidth]
        {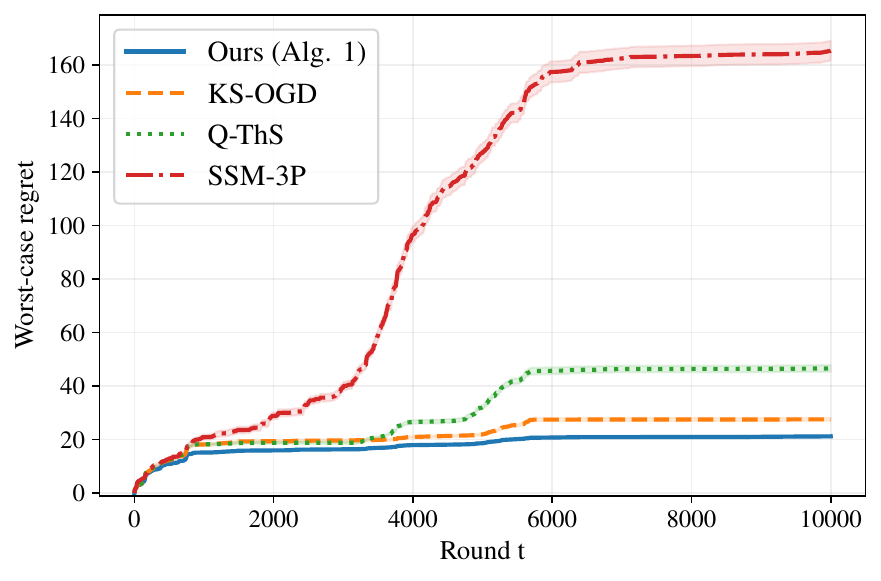}
        \caption{Worst-case regret in adversarial regime.}
        \label{fig:sf ad worst}
    \end{subfigure}
    \hfil
        \begin{subfigure}[t]{0.43\textwidth}
        \centering
        \includegraphics[width=\linewidth]
        {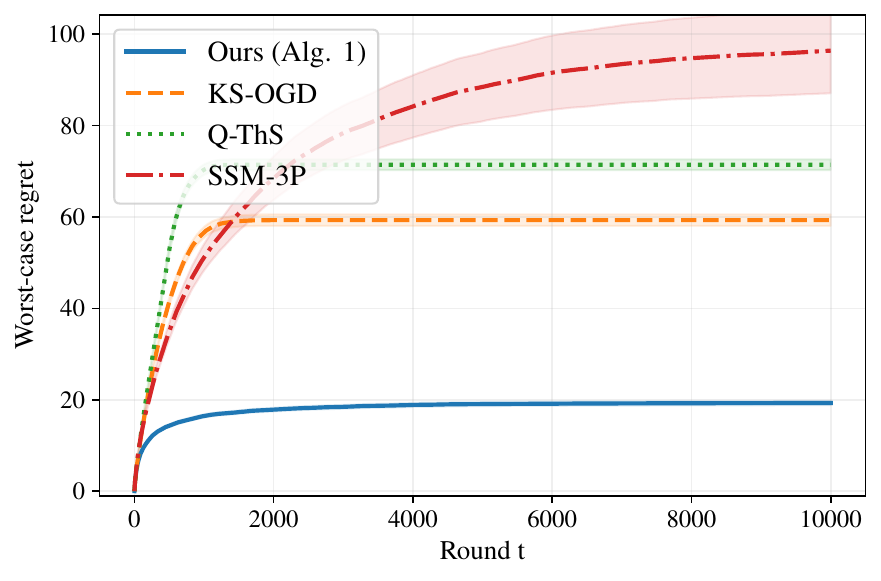}
        \caption{Worst-case regret in stochastic regime.}
        \label{fig:sf st worst}
    \end{subfigure}
    \\
        \begin{subfigure}[t]{0.43\textwidth}
        \centering
        \includegraphics[width=\linewidth]
        {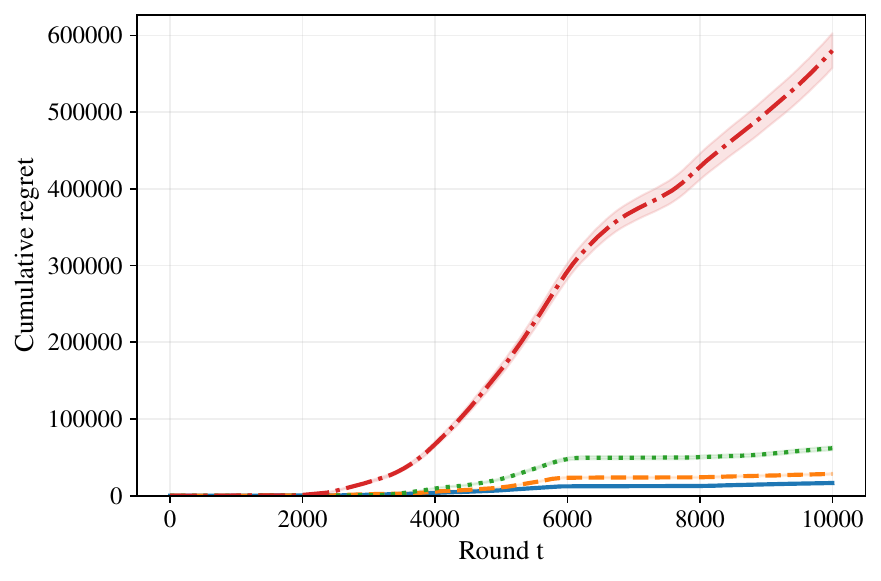}
        \caption{Cumulative regret in adversarial regime.}
        \label{fig:sf ad cum}
    \end{subfigure}
    \hfil
    \begin{subfigure}[t]{0.43\textwidth}
        \centering
        \includegraphics[width=\linewidth]
        {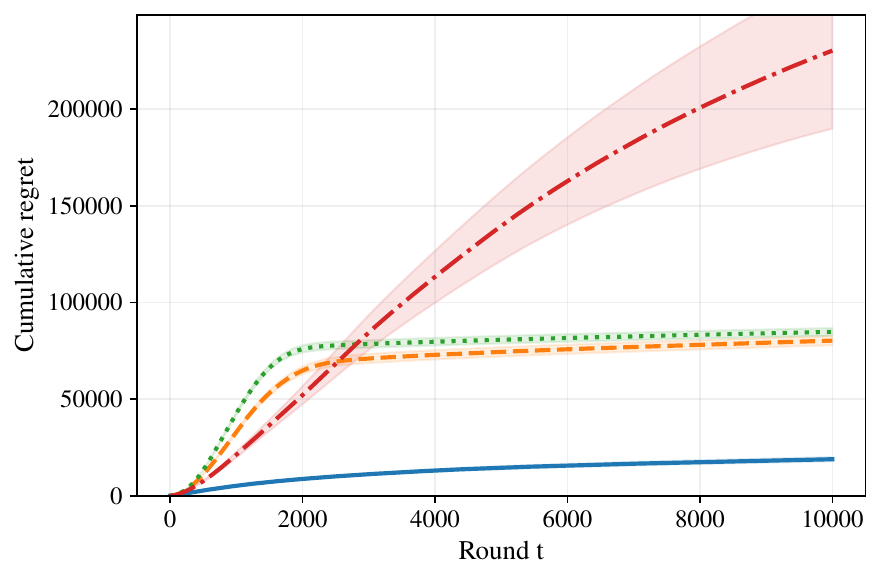}
        \caption{Cumulative regret in stochastic regime.}
        \label{fig:sf st cum}
    \end{subfigure}
    \\
        \begin{subfigure}[t]{0.43\textwidth}
        \centering
        \includegraphics[width=\linewidth]
        {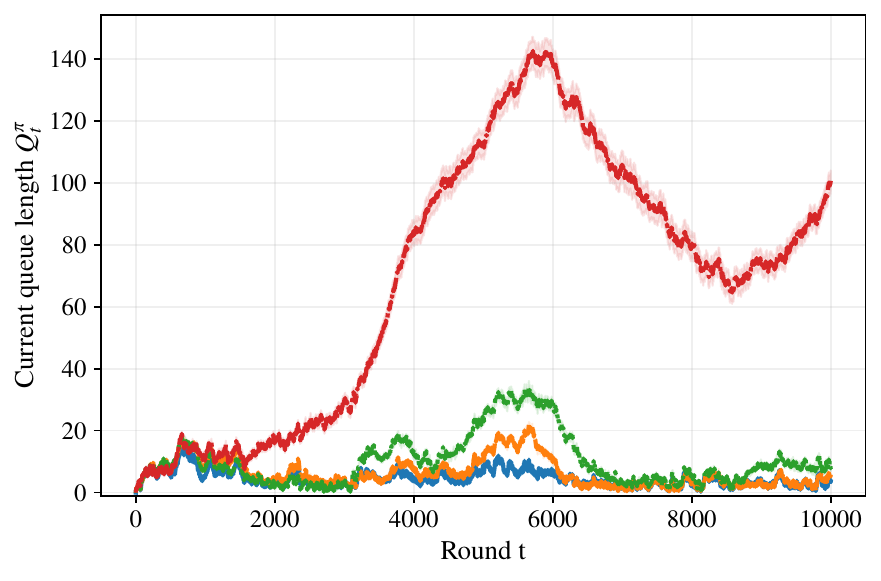}
        \caption{Queue lengths $Q_t^\pi$ in adversarial regime.}
        \label{fig:sf ad queue}
    \end{subfigure}
    \hfil
    \begin{subfigure}[t]{0.43\textwidth}
        \centering
        \includegraphics[width=\linewidth]
        {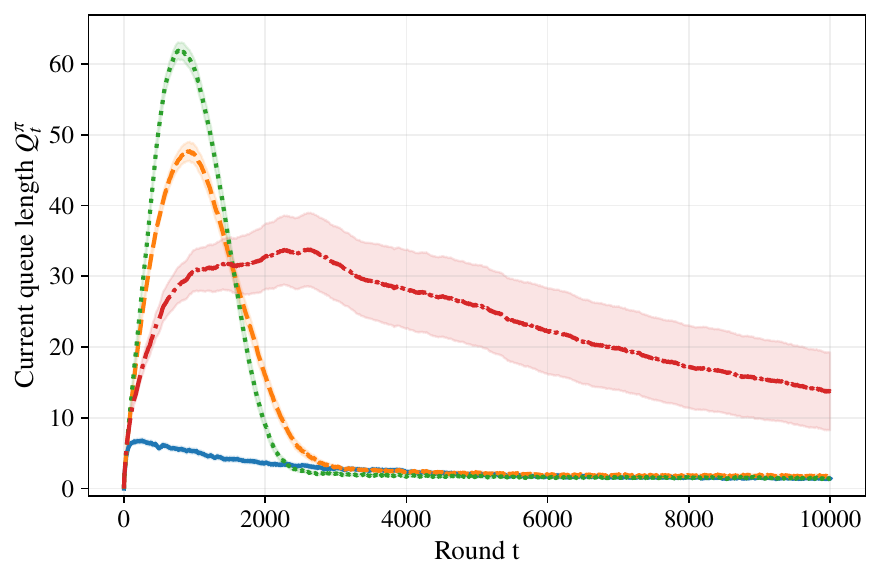}
        \caption{Queue lengths $Q_t^\pi$ in stochastic regime.}
        \label{fig:sf st queue}
    \end{subfigure}
    \caption{Results under service-first dynamics in (\ref{def: queue stan})} \vspace{-0.5em}
    \label{fig: sf rslt}
\end{figure*}

To address this difference, Figures~\ref{fig: af rslt} and~\ref{fig: sf rslt} provide additional results in which all policies are run under the arrivals-first dynamics in~(\ref{def: queue lindley}) and under the service-first in (\ref{def: queue stan}), respectively.
These figures show that the overall qualitative trends remain largely unchanged.

All baseline policies and parameter settings remain unchanged, while our policy is modified only as described below under the service-first convention.
As discussed in Remark~1, under the service-first convention, our policy does not need to observe $A_t$ at the beginning of each round. 
Accordingly, we define an empty period by $Q_t^\pi=0$, matching the definition used by the baseline policies.

We implement Q-ThS, SSM-3P, and KS-OGD by translating the MATLAB code released by \citet{krishnakumar2025minimizing} into Python while preserving its algorithmic update rules and parameter settings.
The code and instructions required to reproduce all experiments will be made publicly available on GitHub.

\end{document}